\documentclass{article}

\usepackage{microtype}
\usepackage{graphicx}
\usepackage{booktabs} 
\usepackage{tabularx}
\usepackage{caption}
\usepackage{subcaption}
\usepackage{multirow}
\usepackage{tikz}
\usepackage{amsmath}
\usepackage[skip=0cm,list=true]{subcaption}
\usepackage{float} 
\usepackage{pifont}
\usepackage{arydshln} 
\usepackage{url}

\usepackage{hyperref}

\usepackage[accepted]{mlsys2025}

\mlsystitlerunning{FLStore: Efficient Federated Learning Storage for non-training workloads}

\begin{document}

\twocolumn[
\mlsystitle{FLStore: Efficient Federated Learning Storage for non-training workloads}

\mlsyssetsymbol{equal}{*}

\begin{mlsysauthorlist}
\mlsysauthor{Ahmad Faraz Khan}{equal,vt}
\mlsysauthor{Samuel Fountain}{equal,umn}
\mlsysauthor{Ahmed M. Abdelmoniem}{qm}
\mlsysauthor{Ali R. Butt}{vt}
\mlsysauthor{Ali Anwar}{umn}
\end{mlsysauthorlist}

\mlsysaffiliation{vt}{Department of Computer Science, Virginia Tech, Blacksburg, USA}
\mlsysaffiliation{umn}{Department of Computer Science, University of Minnesota, Minnesota, USA}
\mlsysaffiliation{qm}{School of Electronic Engineering \& Computer Science, Queen Mary University of London, London, United Kingdom}

\mlsyscorrespondingauthor{Ahmad Faraz Khan}{ahmadfk@vt.edu}

\mlsyskeywords{Machine Learning, Federated Learning, non-training workloads, serverless}

\vskip 0.3in

\begin{abstract}
Federated Learning (FL) is an approach for privacy-preserving Machine Learning (ML), enabling model training across multiple clients without centralized data collection. With an aggregator server coordinating training, aggregating model updates, and storing metadata across rounds. In addition to training, a substantial part of FL systems are the non-training workloads such as scheduling, personalization, clustering, debugging, and incentivization. Most existing systems rely on the aggregator to handle non-training workloads and use cloud services for data storage. This results in high latency and increased costs as non-training workloads rely on large volumes of metadata, including weight parameters from client updates, hyperparameters, and aggregated updates across rounds, making the situation even worse. 
We propose FLStore, a serverless framework for efficient FL non-training workloads and storage. FLStore unifies the data and compute planes on a serverless cache, enabling locality-aware execution via tailored caching policies to reduce latency and costs. Per our evaluations, compared to cloud object store based aggregator server FLStore reduces per request average latency by $71\%$ and costs by $92.45\%$, with peak improvements of $99.7\%$ and $98.8\%$, respectively. Compared to an in-memory cloud cache based aggregator server, FLStore reduces average latency by $64.6\%$ and costs by $98.83\%$, with peak improvements of $98.8\%$ and $99.6\%$, respectively. FLStore integrates seamlessly with existing FL frameworks with minimal modifications, while also being fault-tolerant and highly scalable.
\end{abstract}
]
\printAffiliationsAndNotice{\mlsysEqualContribution} 

\newif\ifsubmission
\submissionfalse

\ifsubmission
\newcommand{\ahmed}[1]{}
\newcommand{\arb}[1]{}
\newcommand{\faraz}[1]{}
\newcommand{\anwar}[1]{}
\newcommand{\fountain}[1]{}
\newcommand{\redwan}[1]{}

\else
\newcommand{\arb}[1]{\textit{\textcolor{red}{{ARB:}#1}}}
\newcommand{\ahmed}[1]{\textit{\textcolor{green}{{Ahmed:}#1}}}
\newcommand{\faraz}[1]{\textit{\textcolor{orange}{{Faraz:}#1}}}
\newcommand{\new}{\textcolor{blue}}
\renewcommand{\new}[1]{#1}
\newcommand{\ali}[1]{\textit{\textcolor{brown}{{Anwar:}#1}}}
\newcommand{\redwan}[1]{\textit{\textcolor{magenta}{{Redwan:}#1}}}
\newcommand{\fountain}[1]{\textit{\textcolor{teal}{{Fountain:}#1}}}

\fi





\section{Introduction}
\label{sec:Introduction}

Federated Learning (FL)~\cite{mcmahan2017communication} is as a privacy-aware solution for ML training across numerous clients without data centralization.
The FL process also encompasses a broad range of non-training workloads. \textit{Non-training workloads} refer to tasks such as scheduling~\cite{lai2021oort, refl}, personalization~\cite{NEURIPS2020ifca,tan2022towards}, clustering~\cite{liu2022auxo}, debugging~\cite{fedDebug}, and incentivization~\cite{tiff,incentive_TPDS}, etc. that are necessary for the success and efficiency of the FL process.
The growing interest in Explainable AI ~\cite{GadeExplainableAI,MohseniExplainableA}, has led to several Explainable FL (XFL) systems that depend on non-training workloads including debugging~\cite{FedDNNDebugger,fedDebug}, accountability~\cite{balta2021accountable,baracaldo2022accountable,YangIoTaccountableFL}, transparency~\cite{tiff}, and reproducibility~\cite{BlockFLAaccountableFL,fedDebug}. 

\begin{figure}[t]
  \centering
  \includegraphics[width=\linewidth]{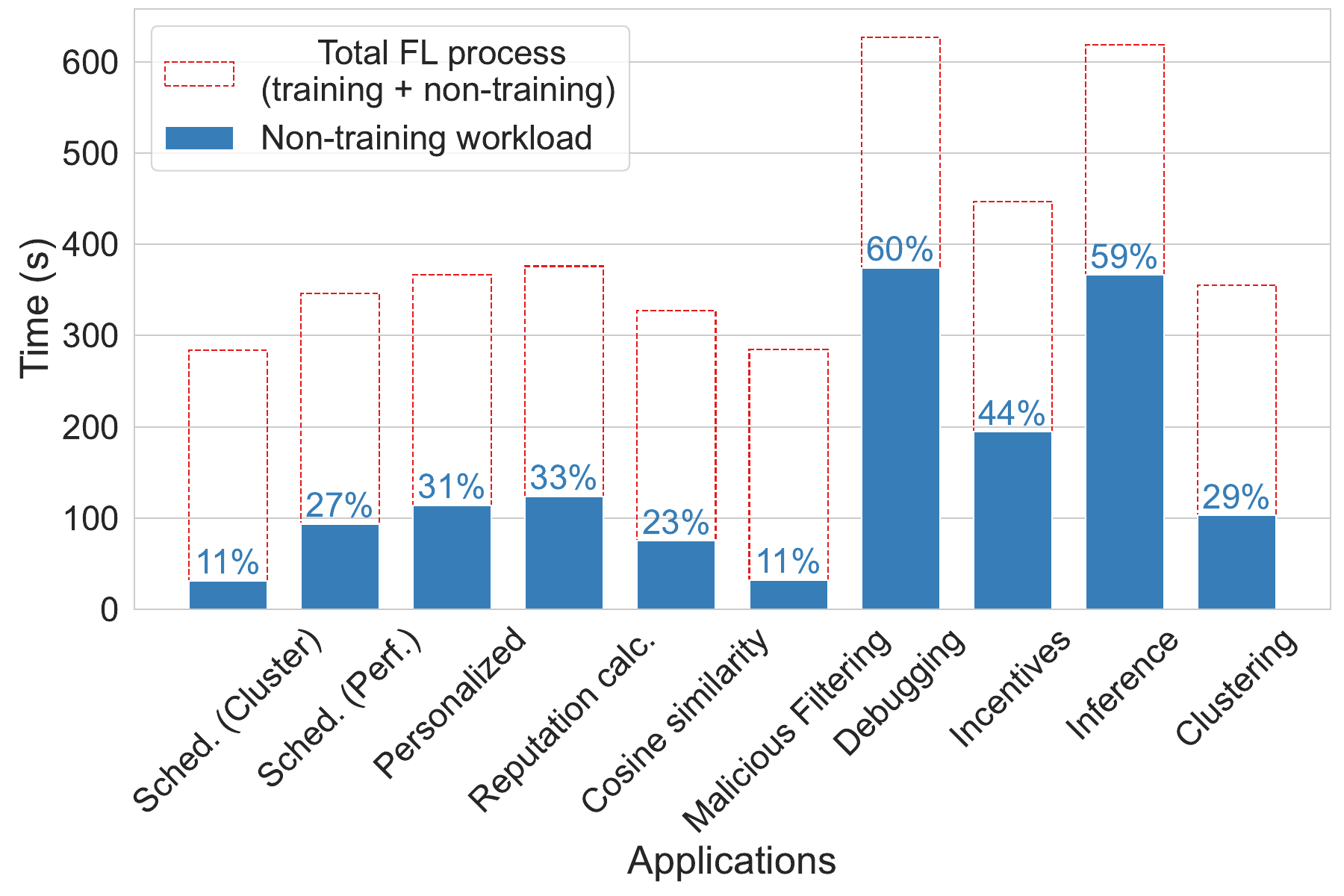}
  \vspace{-10pt}
  \caption{Non-training portion of latency in total FL process per round with 200 clients, EfficientNet model~\cite{tan2021efficientnetv2}, 1000 training rounds, and CIFAR10 Dataset~\cite{krizhevsky2009learning}.}
  \label{fig:Vanilla_FL_vs_non_training_workloads}
\end{figure}

\vspace{-10pt}
\paragraph{Challenges} Existing research concentrates only on training efficiency
~\cite{pmlr-v108-reisizadeh20a_quantization,UVeQFed_quant,yu2023heterogeneous_pruning,kairouz2019advances,FL_survey,lai2021oort, towards_pFL}.
However, non-training workloads constitute a significant and equally important part of the latency and cost in the FL process~\cite{kairouz2019advances}. 
Figure~\ref{fig:Vanilla_FL_vs_non_training_workloads} shows a single non-training application can comprise up to $60\%$ of the total latency of the FL job, and several non-training applications are often executed in the same FL process~\cite{baracaldo2022accountable} with latency several times more than training (\S~\ref{subsection:Non-training workloads in FL}). Non-training workloads are highly data intensive and require tracking, storage, and processing of data, including model parameters, training outcomes, hyperparameters, and datasets reaching thousands of TBs across just 100 FL jobs (\S~\ref{subsection: Shortcomings of popular FL frameworks}). 

\begin{figure}[t]
  \centering
\includegraphics[width=1\columnwidth]{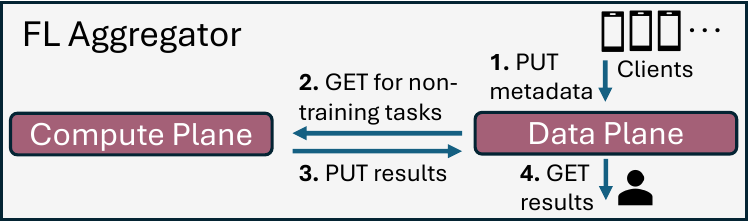}
  \caption{Data flow of serving non-training requests in conventional FL aggregators}
  \label{fig:FLStore_architecture_comparison}
\end{figure}

In current state-of-the-art FL frameworks~\cite{LiFL,bonawitz2019towards,beutel2020flower,fedML,ibmflgithub,federatedai_fate}, cloud-based aggregators handle the non-training workloads and utilize a separate cloud object store for data storage~\cite{aws_s3_api_2024} 
as shown in Figure~\ref{fig:FLStore_architecture_comparison}. 
Consequently, aggregators are ill-equipped to store and process large volumes of FL metadata efficiently and cost-effectively. 

\begin{figure}[t]
  \centering
  \includegraphics[width=\linewidth]{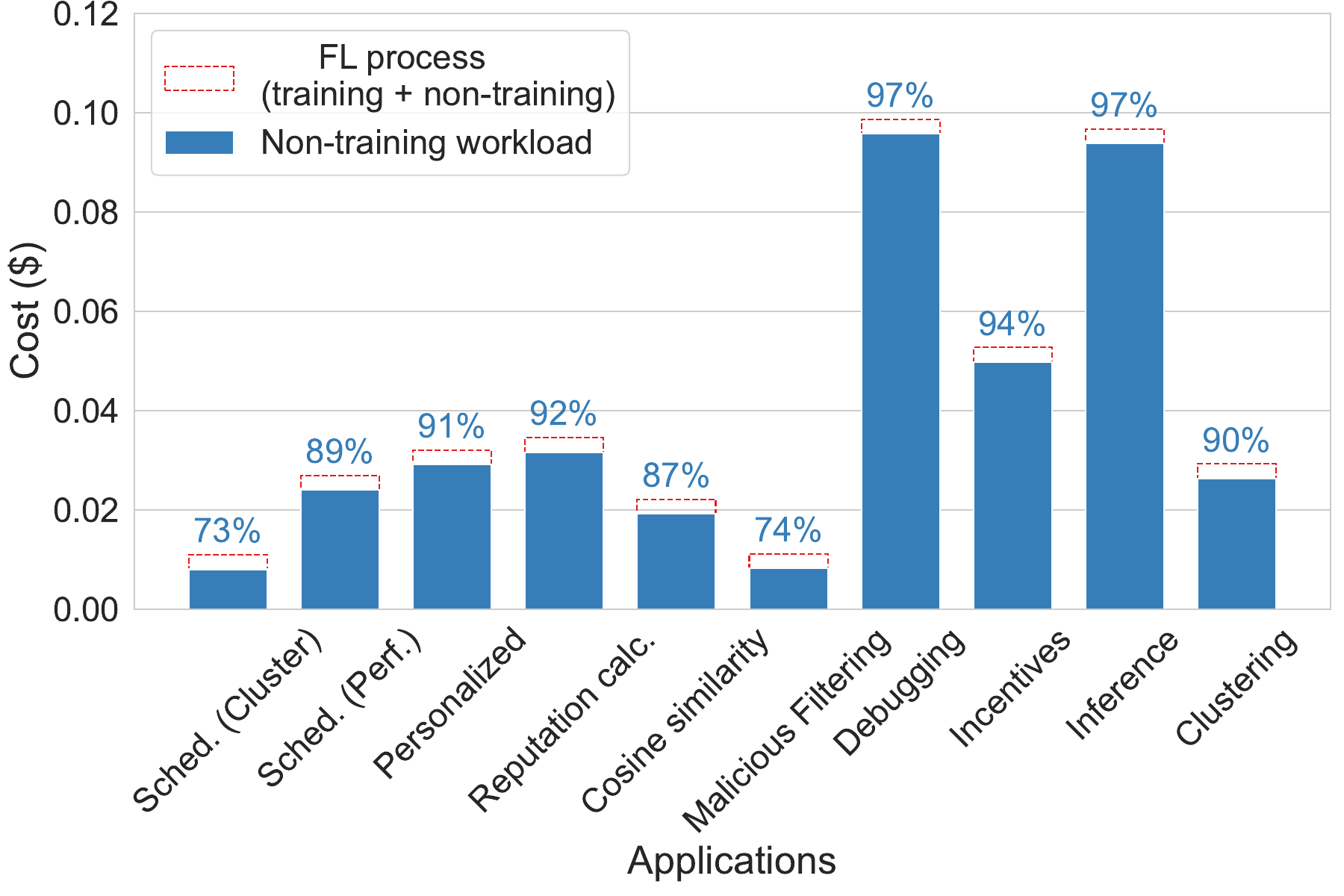}
  \vspace{-10pt}
  \caption{Non-training portion of cost in total FL process per round with 200 clients, EfficientNet model~\cite{tan2021efficientnetv2}, 1000 training rounds, and CIFAR10 Dataset~\cite{krizhevsky2009learning}.}
  \label{fig:Vanilla_FL_vs_non_training_workloads_cost}
\end{figure}

This raises several challenges regarding costs and latency. First, utilizing cloud object stores for data storage separates the data and compute planes. As shown in Figure~\ref{fig:FLStore_architecture_comparison}, this results in extra round trips of storing and fetching the data into the aggregator server's memory, leading to high latency and costs. Even when augmented with more expensive cloud-based caches~\cite{aws_elasticache} the communication bottleneck
remains a challenge~\cite{cacheEnabledFL_Liu}. Second, non-training workloads in FL have diverse data storage and processing requirements. 
For instance, tracing the provenance of specific clients necessitates access to client model updates from previous training rounds~\cite{baracaldo2022accountable}, while identifying issues in malicious clients requires the model updates of all clients for a specific training round~\cite{fedDebug}. Thus, any caching solution for non-training workloads with traditional caching policies that do not consider these unique data requirements will result in sub-optimal performance. 

Third, relying on dedicated servers for executing these workloads becomes a significant issue since the demand for non-training tasks such as debugging and auditing could extend beyond the training phase, necessitating continuous operation of the servers and cache~\cite{baracaldo2022accountable}. 

\vspace{-10pt}
\paragraph{Our Solution}

To address these challenges, we make three key observations. First, unifying the compute and data planes can significantly reduce communication bottlenecks. Second, the iterative nature of FL leads to non-training workloads having sequential and predictable data access patterns; for example, tracking a client's model updates across training rounds will require repeated access to the same client's data across rounds. Third, because non-training workloads, such as debugging, may be required long after training has concluded, a scalable and on-demand solution is essential.

We present FLStore, a caching framework that unifies the data and compute planes with a cache built on serverless functions. FLStore utilizes the co-located compute available on those functions for locality-aware execution of non-training workloads.
FLStore uniquely leverages the iterative nature of FL and its sequential data access patterns to implement tailored caching policies optimized for FL. To develop these policies, we classify non-training workloads in FL applications into a comprehensive taxonomy, categorizing them by their distinct data needs and access patterns. FLStore then customizes its caching policies to the specific type of non-training request encountered. 

\vspace{-10pt}
\paragraph{Contributions}
Our contributions in this work are as follows: 1)~To the best of our knowledge, we present the first comprehensive study of storage and execution requirements of non-training workloads in FL, analyzing their impact on cost and efficiency. 2) Based on the insights from this study, we identify iterative data access patterns in FL, which we leverage to develop FLStore, a novel caching framework with tailored caching policies that use prefetching for locality-aware execution of FL workloads. FLStore is the first FL framework that unifies the data and compute planes and has native support for non-training FL workloads; 3) FLStore provides a highly scalable solution with its serverless functionality~\cite{infinicache} to meet the demands of serving up to millions of clients in FL~\cite{KhanTowards, kairouz2019advances};
It has a modular design~\cite{abadi2016tensorflow,ludwig2020ibm,refl} and can be integrated into any FL framework with minor modifications. 
4) Compared to state-of-the-art FL frameworks~\cite{ibmflgithubcont,federatedai_fate,beutel2020flower} that are based on cloud services~\cite{aws_elasticache,aws_s3_api_2024,aws_sagemaker,google_cloud_storage_docs}, FLStore 
reduces the average per-request latency by $50.8\%$ and up to $99.7\%$,
and the average costs by $88.2\%$ and up to $98.8\%$.
\section{Background and Motivation}
\label{sec:Preliminaries}

\subsection{Non-training workloads in FL}
\label{subsection:Non-training workloads in FL}


XFL aims to improve FL by addressing issues such as clients submitting flawed models due to data quality problems or sabotage~\cite{tiff,fedDebug}. It also emphasizes auditing and regulatory compliance, especially in collaborations involving diverse entities~\cite{balta2021accountable,YangIoTaccountableFL,baracaldo2022accountable}. FLDebugger~\cite{FLDebugger} assesses the influence of each client's data on global model loss, identifying and correcting harmful clients. FedDebug~\cite{fedDebug} improves reproducibility by enabling the FL process to pause or rewind to specific breakpoints and helps detect malicious clients through differential neuron activation testing.

\vspace{-10pt}
\paragraph{Other FL applications}
Due to the distributed nature of FL, many applications involve non-training tasks like clustering~\cite{liu2022auxo,Duan2021FedGroupAF}, personalization~\cite{pflEmpirical,Ruan2022FedSoftSC,Tang2021PersonalizedFL}, and asynchronous learning~\cite{fedbuff}, which are essential for managing and optimizing the FL process. For example, clustering evaluates client models based on factors like training duration, networks, or energy use~\cite{dynamicFL,liu2022auxo,chai2021fedat}, while personalization groups clients by model parameters, efficiency, or accuracy on held-out data~\cite{Ruan2022FedSoftSC,Tang2021PersonalizedFL}. Incentive mechanisms assess client contributions and reputations via accuracy or Shapley Values~\cite{ipfl,shapleyFL,ICL,incentive_TPDS}, and intelligent client selection relies on analyzing client availability, participation, and performance~\cite{refl,Heterogenity_aware_sched,lai2021oort}. Non-training tasks like debugging and hyperparameter tracking are also crucial for optimizing FL~\cite{fedDebug,FedDNNDebugger}.

Non-training tasks can make up to $60\%$ of the FL workflow, as shown in Figure~\ref{fig:Vanilla_FL_vs_non_training_workloads}. Typically, the FL process incorporates numerous non-training tasks. In this scenario involving multiple tasks such as filtering, scheduling, reputation calculation, incentive distribution, debugging, and personalization, non-training tasks account for $86\%$ of total FL time, lasting $6\times$ longer than training. Figure~\ref{fig:Vanilla_FL_vs_non_training_workloads_cost} summarizes the cost breakdown for various FL tasks. The non-training components dominate the overall cost of the FL process as non-training tasks require special provisioning of cloud services and high data transfer costs. For instance, tasks like debugging, inference, and reputation calculation incur non-training costs that constitute over 90\% of their total costs. In our setup with 200 clients per round and only 10 selected for training—the non-training overhead can reach up to 97\%.
This high proportion indicates that even if training is inherently computationally intensive, the cumulative cost of non-training operations such as filtering, scheduling, and incentive management becomes a significant factor in the overall efficiency of FL systems. Optimizing these non-training workloads is therefore critical to reducing latency and improving cost-effectiveness in FL deployments.

\subsection{Shortcomings of popular FL frameworks}
\label{subsection: Shortcomings of popular FL frameworks}
State-of-the-art FL frameworks, as depicted in Figure~\ref{fig:FLStore_architecture_comparison}, generally utilize an aggregator server on a stateful~\cite{lai2021fedscale,beutel2020flower,ibmflgithub,fedML} or serverless compute plane~\cite{LiFL,serverlessML,fedlessServerless}. The serverless model~\cite{jonas2019cloudprogrammingsimplifiedberkeley} allows cloud providers~\cite{aws_lambda, serverlessML} to manage scaling and maintenance by executing functions on demand, with costs based on usage. However, FL data demands can escalate rapidly, reaching over 1500 TB for 100 training sessions with 10 clients each on the CIFAR10 dataset~\cite{krizhevsky2009learning}. To manage this, the compute plane is connected to a separate data plane using cloud caches like ElastiCache~\cite{aws_elasticache} or object stores like AWS S3~\cite{aws_s3_api_2024} and Google Cloud Storage~\cite{google_cloud_storage_docs}. This separation increases communication steps for non-training tasks, involving multiple rounds from receiving requests to fetching and processing data, and then storing results back, which, along with dedicated cloud services, leads to ongoing costs even when non-training requests are dormant.
Compared to these FL frameworks~\cite{lai2021fedscale,beutel2020flower,ibmflgithub,fedML}, FLStore serves as a one-stop solution that processes non-training requests directly from the serverless cache, asynchronously fetching missing data from persistent storage when needed. It also utilizes tailored caching policies based on a classification of non-training workloads. The design of FLStore is discussed in detail later (\S~\ref{sec:Design}).

\begin{figure}
    \centering
{\includegraphics[width=0.5\textwidth]{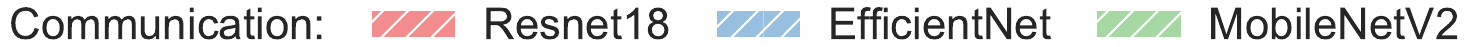}}
\vspace{-0.1em}
{\includegraphics[width=0.5\textwidth]{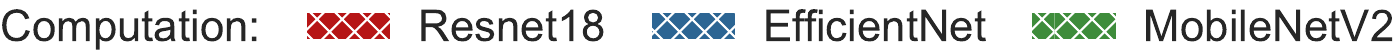}}
\hspace{16em}
\vspace{-0.2em}
\includegraphics[width=\linewidth]{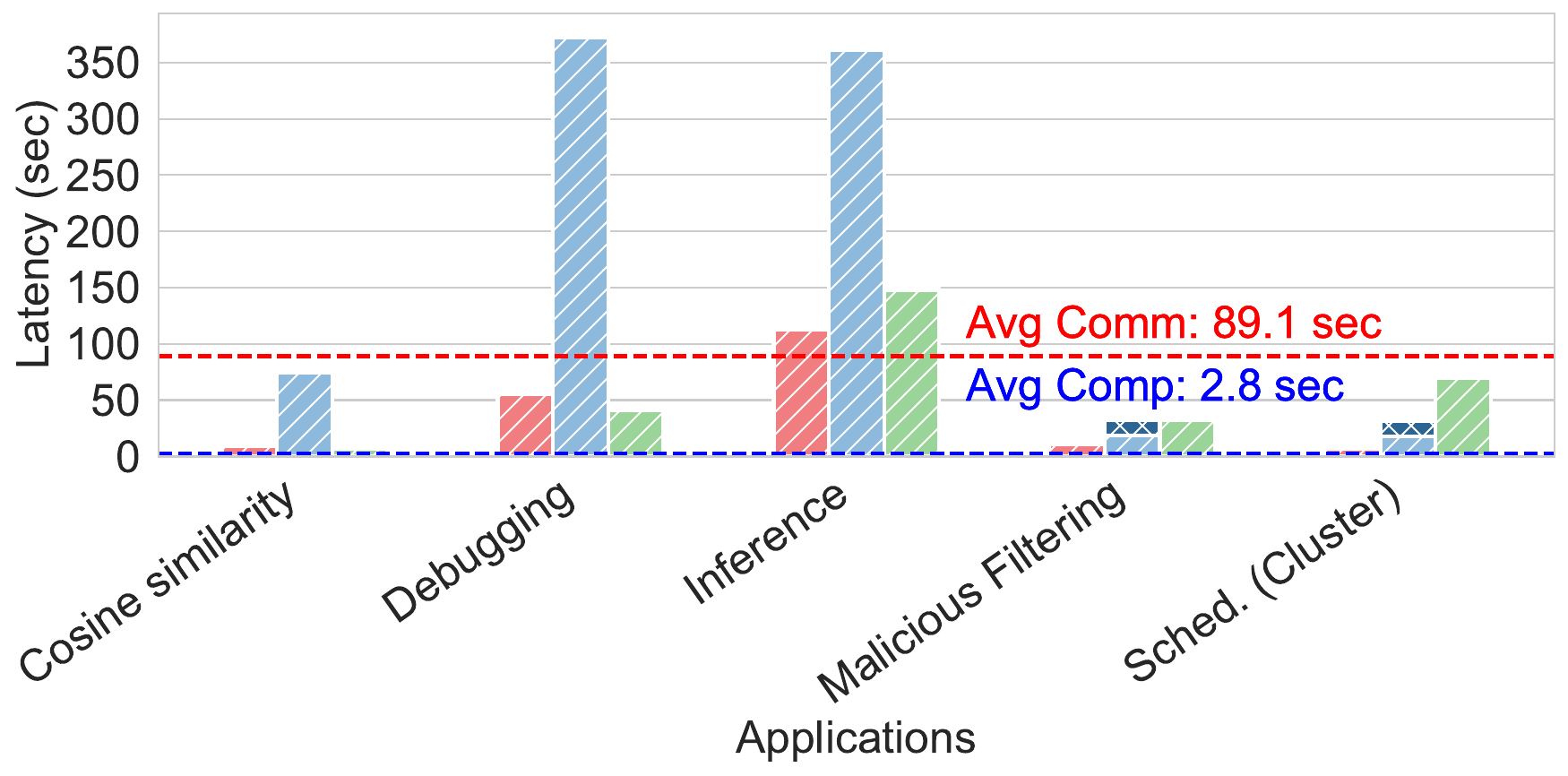}
  \caption{Average workload latencies computation and communication of non-training FL workloads.}
  \label{fig:model_latency_footprint_per_request_workload_latencies}
\end{figure}

\subsection{Serverless Cache for Non-Training Apps}
\label{subsec: Building a Locality-aware Cache on Cloud Functions: Opportunities and Challenges}
To build an in-memory locality-aware cache for non-training FL workloads, we must first answer two important questions: 1) \emph{Can the models utilized in cross-device FL be stored in cloud functions' memory?} 2) \emph{Does the execution latency of non-training workloads fall within the cloud functions lifetime thresholds?} To answer these questions, we first analyze 23 popular models used in cross-device FL settings from various works in FL~\cite{caldas2018leaf,chen2022pflbench,lai2021oort,kairouz2019advances}. 
Analyzing the memory footprint of these models, the average size of these models is approximately $161$ MB as discussed in detail in the Appendix~\ref{sec:Discussion}. These model sizes are perfect for storage in the in-memory cache of cloud functions, as the memory of these functions goes up to 10 GB. We also analyze the typical latency of different non-training workloads. Figure~\ref{fig:model_latency_footprint_per_request_workload_latencies} shows the latencies of executing five different workloads across three different models (EfficientNetV2 Small~\cite{tan2021efficientnetv2}, Resnet18~\cite{he2016deep}, and MobileNet V3 Small~\cite{torchvision_mobilenetv3_small}) and same setup as Figure~\ref{fig:Vanilla_FL_vs_non_training_workloads} on a serverless cloud function~\cite{aws_lambda} while fetching data from a cloud object store~\cite{aws_s3_api_2024}. It can be observed that the average computation latency across workloads is approximately $2.8$ seconds, which is perfect for cloud functions due to their short lifetimes. The small size of the models and the short execution time of non-training tasks for cross-device FL make the memory and compute resources in serverless functions ideal for processing non-training tasks. However, the major bottleneck comes from the $31\times$ higher average communication latency ($89$ sec). Thus, unifying the compute and data planes can ease this bottleneck, enabling efficient, cost-effective serving of non-training requests.



\section{Related Work}
\label{sec:Related Work}
To our knowledge, no existing FL framework efficiently and cost-effectively processes non-training requests.

\textbf{Generic cloud-based frameworks:} General-purpose XAI cloud solutions like AWS SageMaker~\cite{aws_sagemaker} use dedicated instances such as AWS EC2 with storage options like AWS S3~\cite{aws_s3_api_2024} or ElastiCache~\cite{aws_elasticache}. This setup leads to high costs and decreased efficiency due to separated data storage and compute resources~\cite{KhanTowards}, also lacking tailored caching policies suited for FL's iterative nature.

\vspace{-0.5em}
\textbf{FL frameworks:} 
Existing State-of-the-art FL frameworks ~\cite{ibmflgithub, beutel2020flower, fedML, caldas2018leaf, federatedai_fate} follow a similar architecture, where cloud-hosted aggregator servers with separate persistent storage execute non-training tasks~\cite{KhanTowards,bonawitz2019towards,baracaldo2022accountable}, resulting in increased latency and costs.  
\vspace{-0.5em}
\textbf{Serverless aggregators:} Another line of work focuses only on aggregation via serverless functions~\cite{LiFL,KhanTowards,fedlessServerless}. FLStore can easily incorporate aggregation as one of the application workloads, however, FLStore is more generic and also includes additional non-training workloads for FL. Furthermore, non-training workloads such as debugging and incentivization often extend beyond the training phase, requiring aggregators beyond the training phase increasing costs~\cite{fedDebug,KhanTowards,bonawitz2019towards}.

\vspace{-0.5em}
\textbf{Serverless Storage:}
Serverless storage approaches utilize memory available on serverless functions at no additional cost, such as InfiniStore~\cite{10.14778/3587136.3587139}, a cloud storage service, and InfiniCache~\cite{infinicache}, an object caching system using ephemeral functions. These solutions primarily address storage, often underutilizing the computing resources of serverless functions.
\section{FLStore}
\label{sec:Design}
In this section, we present the detailed design for FLStore derived from the following insights we gather from our preliminary analysis (\S~\ref{sec:Preliminaries}):
\vspace{-0.5em}
\begin{itemize}
    \item \textbf{$I_1$:} Communication latency is the major bottleneck for non-training workloads brought by separate compute and data planes in extant solutions (\S~\ref{subsection:Non-training workloads in FL} $\&$~\ref{subsection: Shortcomings of popular FL frameworks}).
    \item \textbf{$I_2$:} Non-training workloads show iterative data access patterns which can be classified, and leveraged to improve performance via a caching solution (\S~\ref{subsection:Non-training workloads in FL}).
    \item \textbf{$I_3$:} Memory footprint of models typically used in cross-device FL and the average latency of non-training workloads are suitable for the inexpensive on-demand Serverless functions (\S~\ref{subsec: Building a Locality-aware Cache on Cloud Functions: Opportunities and Challenges}).
\end{itemize}
\begin{figure}
  \centering
  \includegraphics[width=\linewidth]{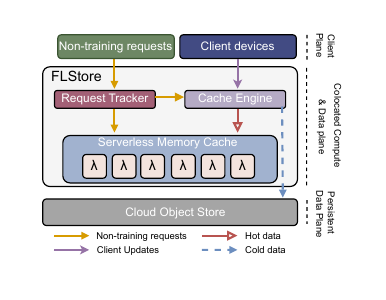}
  \caption{FLStore architecture design.}
  \label{fig:FLStore_architecture}
\end{figure}

\label{subsec:FLStore Cache}

\subsection{Unification of Compute and Data Planes}
\label{subsection:Data tracking over disaggregated memory}
\paragraph{Aims.}
Our first design goal, guided by insight ($I_1$), is to integrate compute and data planes by using serverless function memories for a distributed cache with co-located compute resources like InfiniCache~\cite{infinicache}. However, InfiniCache does not use the compute capabilities of serverless functions or offer specialized caching policies (\S~\ref{subsection: Shortcomings of popular FL frameworks}). This limitation presents a unique opportunity to also utilize free serverless computing for executing non-training workloads ($I_3$).

\paragraph{Challenges.} Creating such a framework presents non-trivial challenges, which we address one by one in the following sections. First, we must track data storage, removal, and updates across multiple function memories (\S~\ref{subsection:Data Tracking Across Serverless Functions}). Second, non-training requests need to be routed to the appropriate functions with the relevant data (\S~\ref{subsection:Locality-aware computation}). Third, it is crucial to identify which metadata should be cached, as storing all metadata would be costly and unsustainable (\S~\ref{subsection:Workload Characterization and Caching}). Lastly, the solution must be scalable, fault-tolerant, and ensure data persistence (\S~\ref{subsec:Fault Tolerance and Data Persistence}). We begin by introducing the main components of our solution (FLStore) that resolve the first challenge of tracking data across functions.

\begin{figure}
  \centering

\begin{subfigure}[t]{\columnwidth}
  \centering
   \includegraphics[width=1\columnwidth]{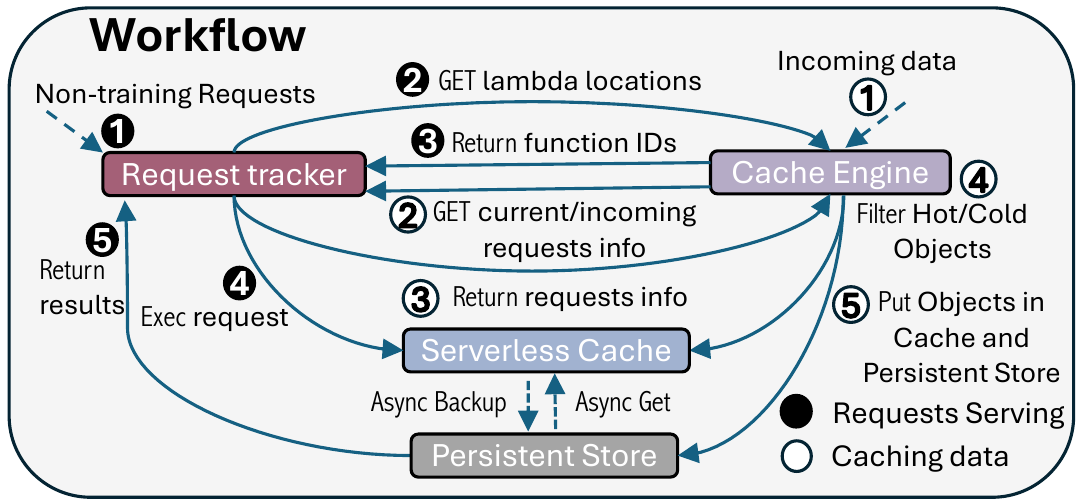}
   \label{subfig:FLStore_workflow}
\end{subfigure}
\begin{subfigure}[t]{\columnwidth}
  \centering
   \includegraphics[width=1\textwidth]{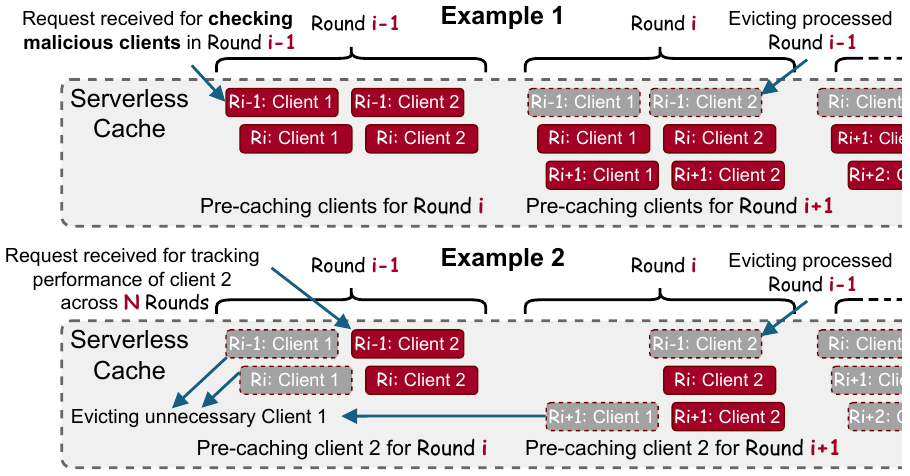}
   \vspace{-10pt}
   \label{subfig:FLStore_workflow_example}
\end{subfigure}
  \caption{FLStore workflow (top) and examples (bottom).}
  \label{fig:FLStore_workflow}
\end{figure}

\vspace{10pt}
\subsection{Tracking Data in Serverless Functions}
\label{subsection:Data Tracking Across Serverless Functions}

FLStore consists of three components, a Request tracker, the Cache Engine, and a Serverless Cache as shown in Figure~\ref{fig:FLStore_architecture}. 
For the Serverless cache, FLStore uses disaggregated serverless function memories similar to~\cite{infinicache}; FLStore extends this design to utilize the serverless compute resources of those functions to process non-training requests. 
The Cache Engine and the Request tracker can be run in the cloud or collocated with a client. The Cache Engine uses a hash table to store the location of data in disaggregated functions, tracking specific metadata to the functions where it is cached. 
The CacheEngine dictionary format is as follows:
\begin{align*}
    Tuple(Client:str, Round:int)\rightarrow FunctionID:str
\end{align*}
 As shown by the data-flow in Figure~\ref{fig:FLStore_workflow}, the Cache Engine receives incoming data from client training devices (Step~\ding{192}) and fetches the current and incoming non-training request information from the Request tracker (Steps~\ding{193} $\&$~\ding{194}). Based on the request types, it utilizes the appropriate caching policy to filter hot data from cold data (Step~\ding{195}) and puts models in Serverless Cache and Persistent Store, respectively (Step~\ding{196}). The data is cached at the granularity of client models such that each function holds at least one client model. This level of granularity is practical as a single function provides up to 10 GB of memory~\cite{aws_lambda}. Unlike conventional cloud caching systems like ElastiCache, FLStore's serverless cache also provides compute resources for non-training tasks, ensuring that cached data is close to the compute needed to execute requests. So next we discuss how to resolve the second challenge of routing the requests to the appropriate functions containing the relevant data for locality-aware execution.


\vspace{10pt}
\subsection{Locality-Aware Request Routing}
\label{subsection:Locality-aware computation}
One of FLStore's key contributions is to effectively leverage local compute resources to process data, enhancing the overall efficiency of resource utilization.
Using these compute resources requires the non-training requests to be routed to the functions with data relevant to the request. 
The Request tracker, as shown in Figure~\ref{fig:FLStore_architecture}, is responsible for receiving requests from clients, forwarding the request to the appropriate functions, and keeping track of the progress. The tracking data is stored in a dictionary where request IDs serve as keys, and the corresponding values include the list of function IDs to which the request was routed and the progress made by each function in executing the request. The Request Tracker dictionary is formatted as follows:
\vspace{-0.5em}
\begin{align*}
    RequestID:str\rightarrow Tuple(&List[FunctionID:str, ...],\\
    &Status:bool)
\end{align*}
Figure~\ref{fig:FLStore_workflow} describes the workflow. 
Upon receiving the request in (Step \ding{202}), the Request tracker fetches the function IDs from the Cache Engine where the data required for the non-training request is cached (Steps \ding{203} and \ding{204}). 
Then, it issues the requests to those function IDs and keeps track of their progress (Step \ding{205}), reporting the results as soon as they are returned to the client daemon (Step \ding{206}).
Next, we discuss how to determine which data is important for caching.

\begin{table}[t]
\centering
\caption{Taxonomy of Non-Training Applications and Mapping of Workloads in FLStore\\}
\label{tab:combined_taxonomy_workloads}
\footnotesize
\setlength{\tabcolsep}{4pt} 
\renewcommand{\arraystretch}{0.7} 
\begin{tabular}{p{0.04\columnwidth} p{0.18\columnwidth} p{0.69\columnwidth}}
\toprule
\textbf{ID} & \textbf{Caching Policy} & \textbf{Applications and Mapped Workloads} \\
\midrule
\textbf{P1} & Individual Client Updates 
& Evaluates individual model's accuracy and fairness~\cite{li2020federated, yu2020fairness, ezzeldin2023fairfed}.\\ 
\addlinespace
\textbf{P2} & All Updates in a Round
& Used in Personalization~\cite{tan2022towards}, Clustering~\cite{NEURIPS2020ifca}, Scheduling~\cite{chai2020tifl}, Contribution calculation~\cite{shapleyFL}, Filtering malicious clients~\cite{tiff}, Cosine Similarity~\cite{liu2022auxo}. \\ 
\addlinespace
\textbf{P3} & Updates Across Rounds 
& Facilitates debugging~\cite{fedDebug, FedDNNDebugger}, fault tolerance~\cite{balta2021accountable,YangIoTaccountableFL}, reproducibility, transparency, data provenance, and lineage~\cite{baracaldo2022accountable}.\\ 
\addlinespace
\textbf{P4} & Metadata $\&$ Hyperparameters 
& Hyperparameter tuning~\cite{zhou2023singleshot}, tracking client resources for scheduling, clustering client priorities~\cite{liu2022auxo}, clustering performance, client incentives, and client dropouts, monitoring payouts~\cite{incentive_TPDS}, and optimizing communication through pruning and quantization~\cite{Khan2024FLOAT,shapleyFL}. \\ 
\bottomrule
\end{tabular}
\end{table}

\vspace{10pt}
\subsection{Workload Characterization and Caching}
\label{subsection:Workload Characterization and Caching}

Based on our insight ($I_2$) from studying existing works~\cite{lai2021oort,beutel2020flower,fedDebug, kairouz2019advances}, we recognize that FL follows an iterative process with sequential data access patterns, which can inform tailored caching policies. We first analyze the data processing needs of popular FL applications to develop a taxonomy of their non-training workloads~\cite{fedDebug, FedDNNDebugger, baracaldo2022accountable,balta2021accountable,tiff} as shown in Table~\ref{tab:combined_taxonomy_workloads}. Leveraging the insights gained from this study, we propose tailored caching policies that also enable easy FLStore extension to new applications.

While a Serverless cache is scalable enough to store all metadata~\cite{zhang2023infinistore}, FL metadata can reach several thousand terabytes (TBs), so using tailored caching policies significantly reduces resource consumption and costs. For example, an FL job with 1000 clients and 1000 training rounds using the EfficientNet model~\cite{tan2021efficientnetv2} would require 79 TBs of memory across 10098 Lambda functions, costing $\$10.2$ per hour or $\$7357.8$ per month. With FLStore's tailored policies, only 1.2 GB is consumed from just two Lambda functions, reducing costs to $\$0.001$ per hour or $\$0.7$ per month.

Table~\ref{tab:combined_taxonomy_workloads} also outlines the corresponding policies for each workload type in the taxonomy. Based on the chosen caching policy, FLStore distinguishes \emph{hot data} from \emph{cold data}, caching the former in serverless memory and asynchronously storing the latter in the persistent store. Next, we discuss each caching policy in detail:


\noindent\textbf{P1: Single Client or Aggregated Model.} This policy applies to tasks such as serving and testing a fully trained model~\cite{li2020federated, yu2020fairness, ezzeldin2023fairfed}, and requires access to individual model updates for fine-tuning~\cite{ijcai2022p311} or the final aggregated model~\cite{SpreadScalableAggregation}. As previously explained (\S~\ref{sec:Preliminaries}), the final aggregated model created by combining updates from participating clients after the FL training concludes is a model ready for deployment to consumers. To support these workloads, this policy requires caching the aggregated model for serving and inference. Additionally, any updates to this model are cached for workloads that involve comparative analysis or tracking of the aggregated model.

\noindent\textbf{P2: All Client Model Updates per Round.} Applications such as filtering malicious clients~\cite{tiff}, calculating clients' relative contributions~\cite{shapleyFL}, debugging~\cite{fedDebug, FedDNNDebugger}, personalization~\cite{towards_personalized}, and fault tolerance~\cite{balta2021accountable} fall under this category because they require iterative access to all client updates for specific rounds. When a request in this category is made for a particular client in a training round, we pre-cache all client updates for that round and the next, as these workloads require iterative access to clients' metadata from the requested round and possibly the next round. Metadata from previous rounds is unnecessary since these applications operate separately and incrementally for each round. Additionally, we keep the latest round cached, as workloads like scheduling, contribution calculation, and malicious client filtering run for each new round, requiring all client updates from that round.\looseness=-1

Figure~\ref{fig:FLStore_workflow} illustrates two example workloads handled by FLStore. The first corresponds to this policy (P2), where a malicious filtering application is executed per round. In this example, data from round \(Ri-1\) is old data that was required for a prior request, while round \(Ri\) was pre-cached during the execution of that prior request. As the current request for round \(Ri\) executes, FLStore evicts past data and pre-caches round \(Ri+1\) for future requests, demonstrating how iterative non-training workloads in FL such as incentive distribution, scheduling, etc. have predictable data needs.

\noindent\textbf{P3: Client Model Updates Across Rounds.} 
Applications like reproducibility, checkpointing, transparency, data provenance, and lineage require access to a single client's model updates across consecutive rounds~\cite{baracaldo2022accountable}. To support these, we cache the client's model update for the requested round and pre-cache that client's metadata from the previous and subsequent rounds. This is necessary because these workloads track performance, costs, or other metrics for a client over time or training rounds.

The second example in Figure~\ref{fig:FLStore_workflow} demonstrates a workflow for this policy (P3). In the example, the system is handling a request to track the improvement of client $2$. Since tracking improvement is an iterative, round-based workload, the cache holds data from round $Ri-1$ (from a past request), while the current request is for round $Ri$, and the next is expected to be for round $Ri+1$. As FLStore processes the current request for round $Ri$, it evicts data from round $Ri-1$ and pre-caches client $2$'s updates for round $Ri+1$. 


\noindent\textbf{P4: Metadata and Hyperparameters.} This includes applications such as hyperparameter tuning~\cite{zhou2023singleshot}, assessing data shift impacts on performance~\cite{NEURIPS2023_data_shift}, tracking client resource availability for scheduling, clustering by client priorities, and monitoring client payouts in FL. Communication optimization techniques like pruning, quantization, and contribution tracking for incentive distribution also require monitoring client optimization and contributions~\cite{Khan2024FLOAT,shapleyFL}.

For these applications, we cache configuration and performance metadata, including hyperparameters, for the most recent $R$ rounds, where $R$ is tunable (default is 10). This ensures that up-to-date data is available for configuration and tuning, as older data may not be reliable. For instance, when scheduling client devices for training, current resource information is critical, as outdated data could cause clients to miss training deadlines.




\textbf{Choice of policy. } Since non-training workloads are iterative with predictable data needs~\cite{baracaldo2022accountable,fedDebug,kairouz2019advances}, we use the mappings in Table~\ref{tab:combined_taxonomy_workloads} to select the appropriate caching policy. While we continue to add new workloads, most fit into existing caching policies due to the iterative nature of FL. Future work includes incorporating a Reinforcement Learning with Human Feedback (RLHF) agent~\cite{Khan2024FLOAT} to adapt policies for outlier workloads. Additional discussion on improving caching policy selection is in the Appendix~\ref{sec:Discussion}.





\begin{figure*}
\centering
\includegraphics[width=0.25\textwidth]{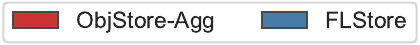}
\vspace{-4pt}

\begin{subfigure}[t]{0.49\textwidth}
  \centering
   \includegraphics[width=1\textwidth]{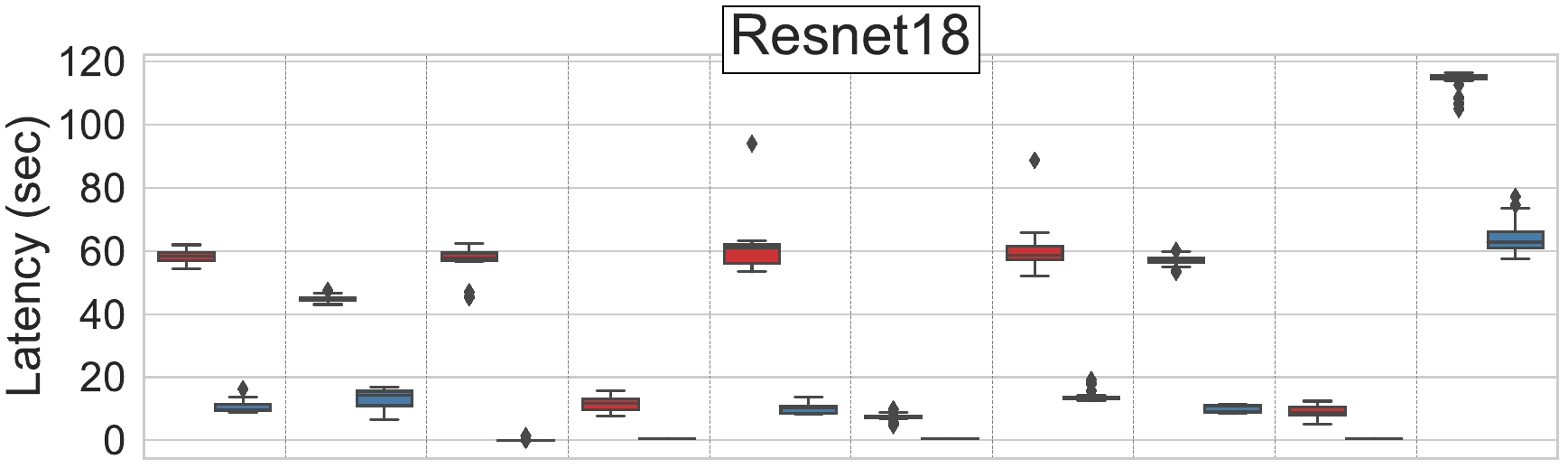}
   \vspace{-10pt}
   \label{subfig:FLStore_total_time_resnet18}
\end{subfigure}
\begin{subfigure}[t]{0.49\textwidth}
  \centering
   \includegraphics[width=1\textwidth]{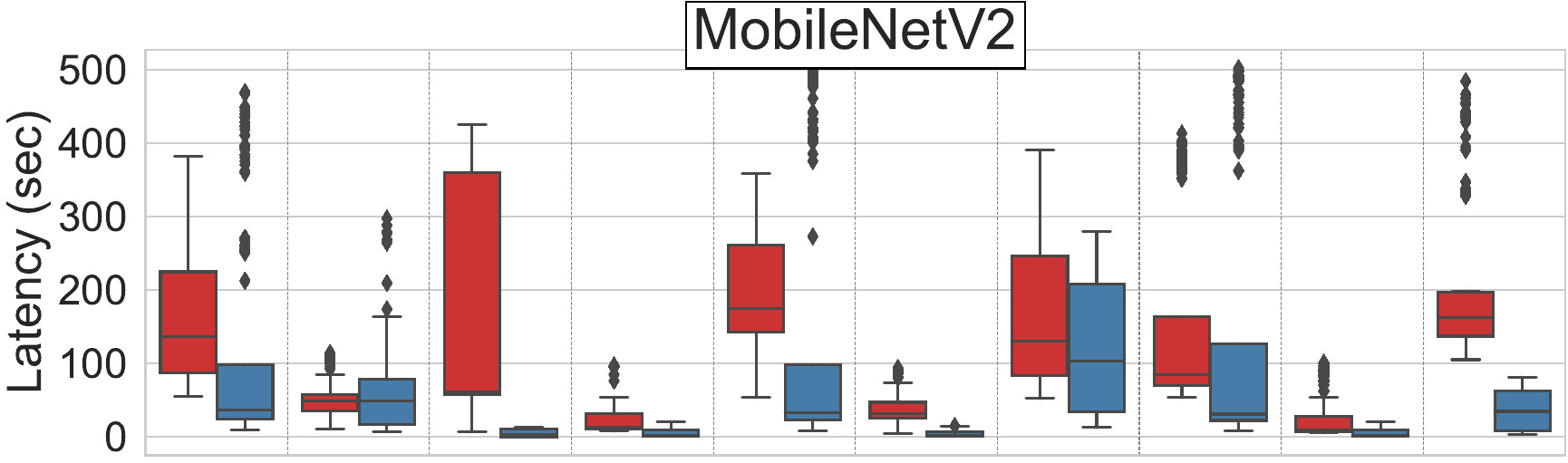}
   \vspace{-10pt}
   \label{subfig:FLStore_total_time_MobileNetV2}
\end{subfigure}
\begin{subfigure}[t]{0.49\textwidth}
  \centering
   \includegraphics[width=1\textwidth]{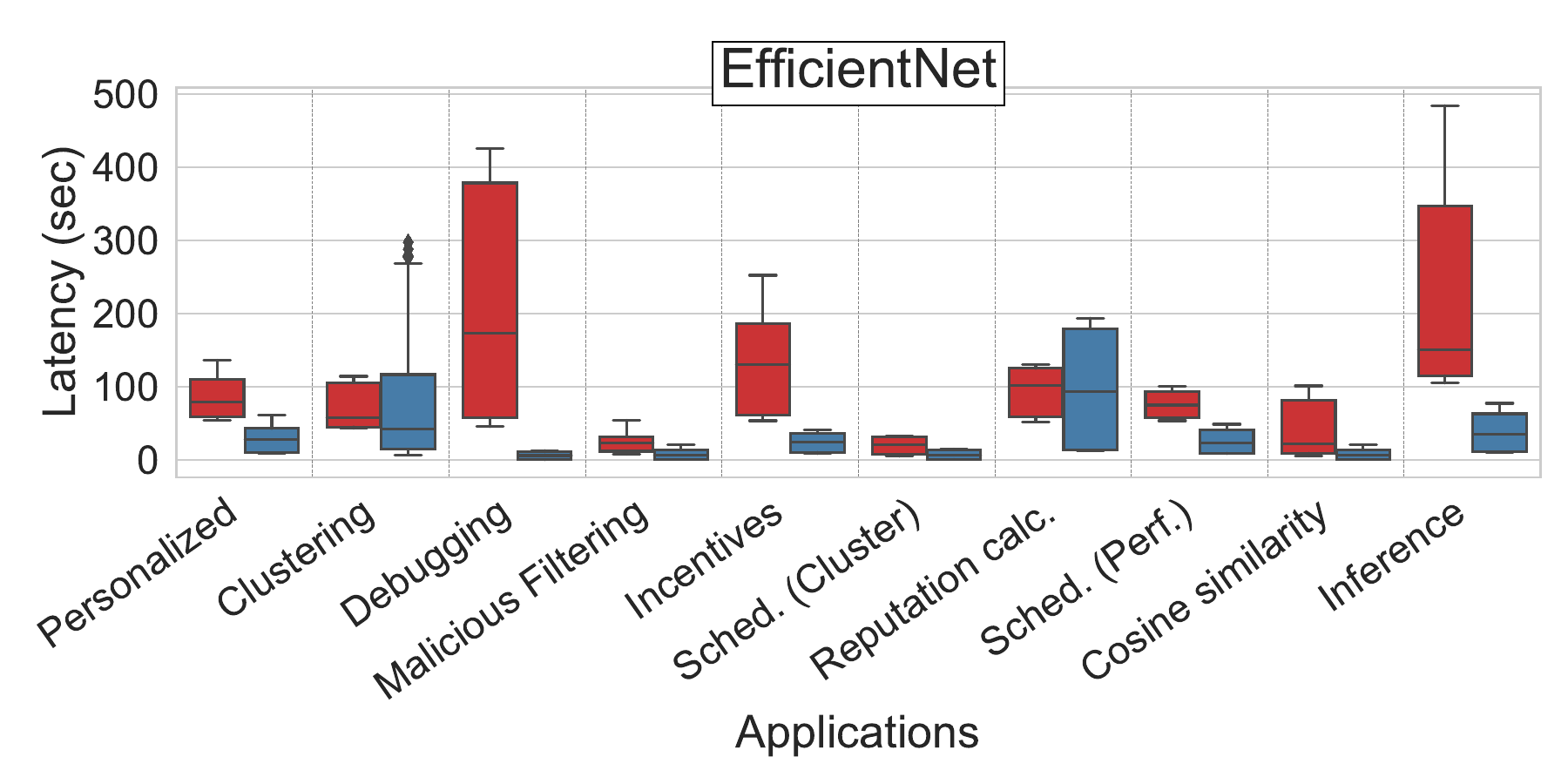}
   \vspace{-10pt}
   \label{subfig:FLStore_total_time_EfficientNet}
\end{subfigure}
\begin{subfigure}[t]{0.49\textwidth}
  \centering
   \includegraphics[width=1\textwidth]{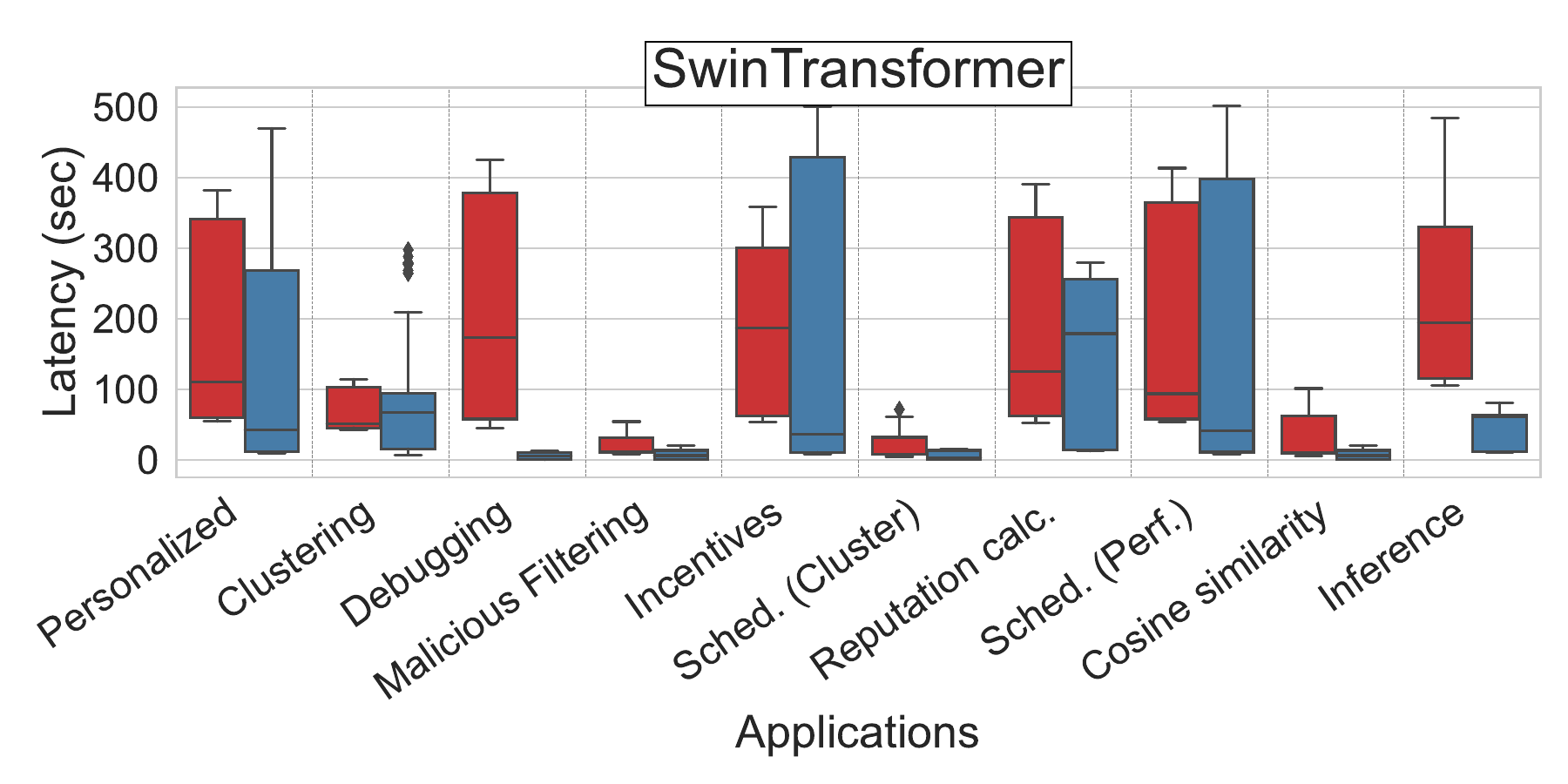}
   \vspace{-10pt}
   \label{subfig:FLStore_total_time_SwinTransformer}
\end{subfigure}
  \caption{FLStore vs. Baseline \textbf{per request latency} comparison over 50 hours.}
  \label{fig:per_request_time_comparison}
\end{figure*}

\vspace{10pt}
\subsection{Data Persistence and Fault Tolerance}

In this section, we discuss how FLStore ensures data persistence and fault tolerance against reclaimed serverless functions. The persistent store serves as a \emph{cold data} repository for all data as protection against data loss and allows users to revisit data from past rounds. This data is crucial for post-training analysis, such as distributing incentives or visualizing convergence and loss trends.  In the rare event that all cached functions fail, FLStore retrieves the necessary data from the persistent store, similar to state-of-the-art FL frameworks~\cite{federatedai_fate,beutel2020flower,ibmflgithub}, ensuring comparable performance.

FLStore addresses fault tolerance through prevention and mitigation. We regularly ping cached functions to check their liveness, leveraging cloud platforms' default behavior~\cite{zhang2023infinistore}. Cloud providers like AWS~\cite{aws_lambda} cache functions at no cost, as long as they are regularly invoked~\cite{zhang2023infinistore}. Pinging a function every minute, as recommended by InfiniStore~\cite{zhang2023infinistore}, incurs a minimal monthly cost of $\$0.0087$ per instance and $\$0.00000016$ per million requests.
Additionally, FLStore replicates functions to enhance reliability. Each primary function has $k$ secondary copies to prevent stragglers and recovery delays. If the primary function fails to acknowledge a request or respond within a set time, the Request Tracker reroutes the request to a secondary instance.
For added reliability, we recommend scaling function instances linearly with the number of requests, which minimizes cost and latency while preventing data re-fetching and cold starts.
\label{subsec:Fault Tolerance and Data Persistence}
\paragraph{Scalability over Serverless Functions}
FLStore’s cache has two scalable facets: the cache size and handling more concurrent requests. To increase cache size, new serverless functions can be spawned to store additional data. For concurrent requests, new functions can be spawned which are simply copies of existing ones. Since serverless functions are highly scalable~\cite{infinicache}, scaling FLStore’s cache is straightforward—new function instances are created as needed. FLStore can also spawn multiple instances to enhance scalability and performance.

\section{Evaluation}
\label{sec:Evaluation}


\subsection{Evaluation Setup}
\label{subsec:Evaluation Setup}
This section presents a proof-of-concept analysis to demonstrate the potential improvements brought by FLStore in latency and cost for non-training FL workloads. We show the effectiveness
of FLStore by answering the questions:
\vspace{-0.7em}
\begin{itemize} 
    \item How well does FLStore reduce the latency of non-training workloads compared to state-of-the-art FL frameworks? (\S~\ref{subsec: Latency Analysis})
    \item How is the performance of FLStore's tailored caching policies compared to traditional ones? (\S~\ref{subsec:FLStore caching policies})
    \item What is the overhead of FLStore components? (\S~\ref{subsec: Cache Engine and Request Tracker Overhead})
    \item How well does FLStore scale for parallel FL jobs? (\S~\ref{subsec:Scalability of FLStore})
    \item How well does FLStore cope with faults? (\S~\ref{subsec:Fault Tolerance})
\end{itemize} 

\vspace{-0.5em}
\paragraph{Baselines:} 
We utilize baselines derived from the architectures of popular FL frameworks~\cite{LiFL,fedML,ibmflgithub,beutel2020flower}, as depicted in Figure~\ref{fig:FLStore_architecture_comparison}. Specifically, we deploy the cloud aggregator server on the $ml.m5.4xlarge$ instance of AWS SageMaker~\cite{aws_sagemaker}, a widely-used AWS service for managing non-training workloads such as inference and debugging~\cite{sagemaker_Liberty,sagemaker_Perrone,sagemaker_Piali}. AWS SageMaker connects with data storage options such as AWS S3~\cite{aws_s3_api_2024} for cloud object storage or AWS ElastiCache~\cite{aws_elasticache} for in-memory caching.
Thus, our baselines are structured as follows: the first features an aggregator server on AWS SageMaker linked with AWS S3 (ObjStore-Agg), and the second connects AWS SageMaker with ElastiCache (Cache-Agg). In both setups, the data plane stores all FL metadata, while AWS SageMaker, forming the compute plane, processes non-training requests.
\vspace{-0.5em}
\paragraph{Workloads:} We evaluate ten common non-training workloads, integral to many FL applications as shown in Table~\ref{tab:combined_taxonomy_workloads}, across four models: EfficientNetV2 Small~\cite{tan2021efficientnetv2}, Resnet18~\cite{he2016deep}, MobileNet V3 Small~\cite{torchvision_mobilenetv3_small}, and SwinTransformerV2 tiny~\cite{liu2021swinv2}. Each model underwent FL training with 10 clients per round, selected from a pool of 250, across 1000 rounds or until convergence, following standard cross-device FL protocols in related studies~\cite{lai2021oort, kairouz2019advances}.
\vspace{-0.5em}
\paragraph{Metrics:} Since throughput can be effectively managed through scaling, we focus on evaluating the latency and cost associated with communication and computation. We assess these metrics per request and their aggregated total for multiple requests over a period of several days, encompassing various non-training workload applications and models.

\paragraph{Implementation of FLStore:} 
FLStore is implemented using the OpenFaas serverless framework~\cite{openfaas_faas}. Function sizes are automatically adjusted to accommodate the varying model sizes, with larger function allocations (2 CPU cores and 4 GB of memory) configured for SwinTransformer and EfficientNet models and 1 CPU core and 2 GB of memory for Resnet 18 and MobileNet models.
For both the baseline and FLStore setups, we use MinIO~\cite{minio} as our persistent data store, which is compatible with Amazon S3~\cite{aws_s3_api_2024}. The MinIO configuration involves a 3-node cluster, with each node hosting six IronWolf 10TB HDDs (7200 RPM) and running default MinIO settings.
\begin{figure*}
  \centering
\includegraphics[width=0.25\textwidth]{figures/Legend_50_hours.pdf}
\vspace{-4pt}


\begin{subfigure}[t]{0.49\textwidth}
  \centering
   \includegraphics[width=1\textwidth]{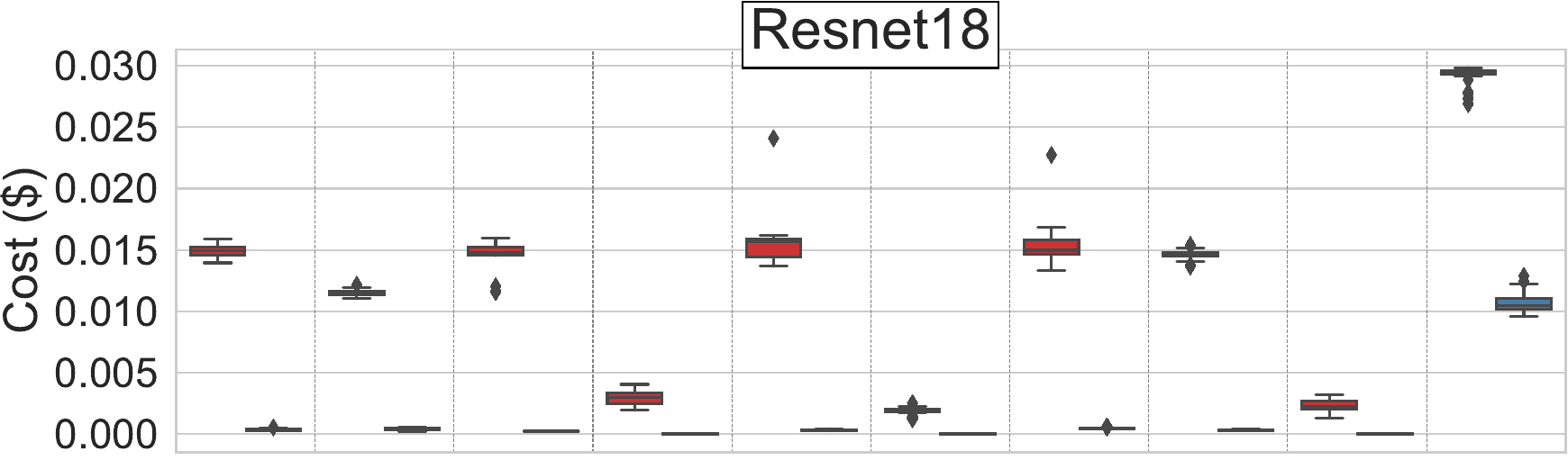}
   \vspace{-10pt}
   \label{subfig:FLStore_total_cost_resnet18}
\end{subfigure}
\begin{subfigure}[t]{0.49\textwidth}
  \centering
   \includegraphics[width=1\textwidth]{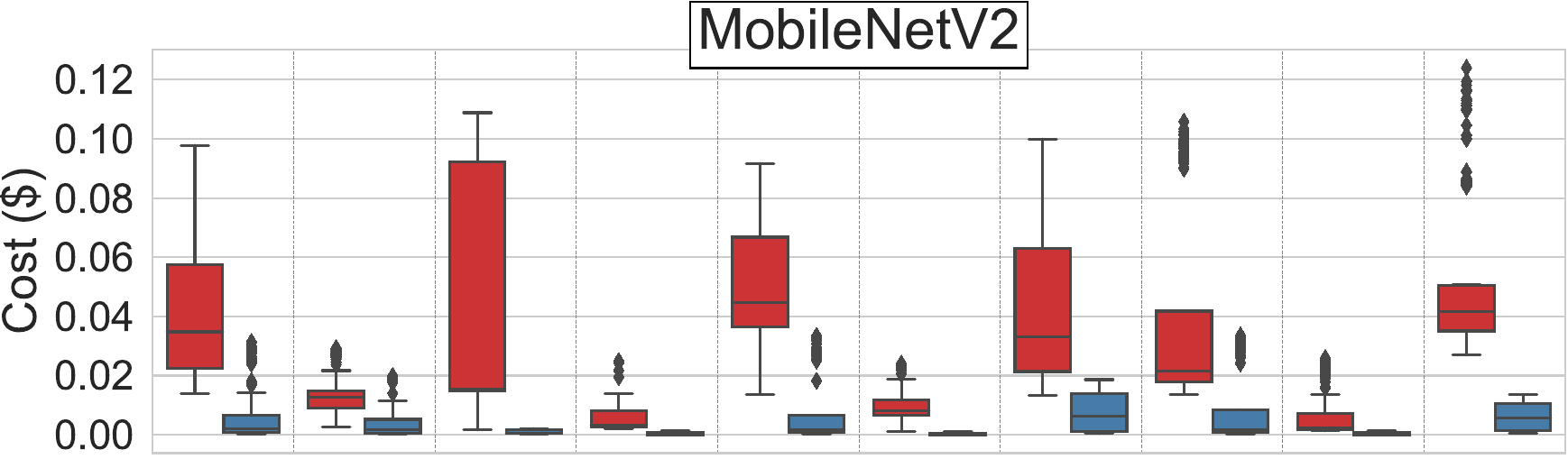}
   \vspace{-10pt}
   \label{subfig:FLStore_total_cost_MobileNetV2}
\end{subfigure}
\begin{subfigure}[t]{0.49\textwidth}
  \centering
   \includegraphics[width=1\textwidth]{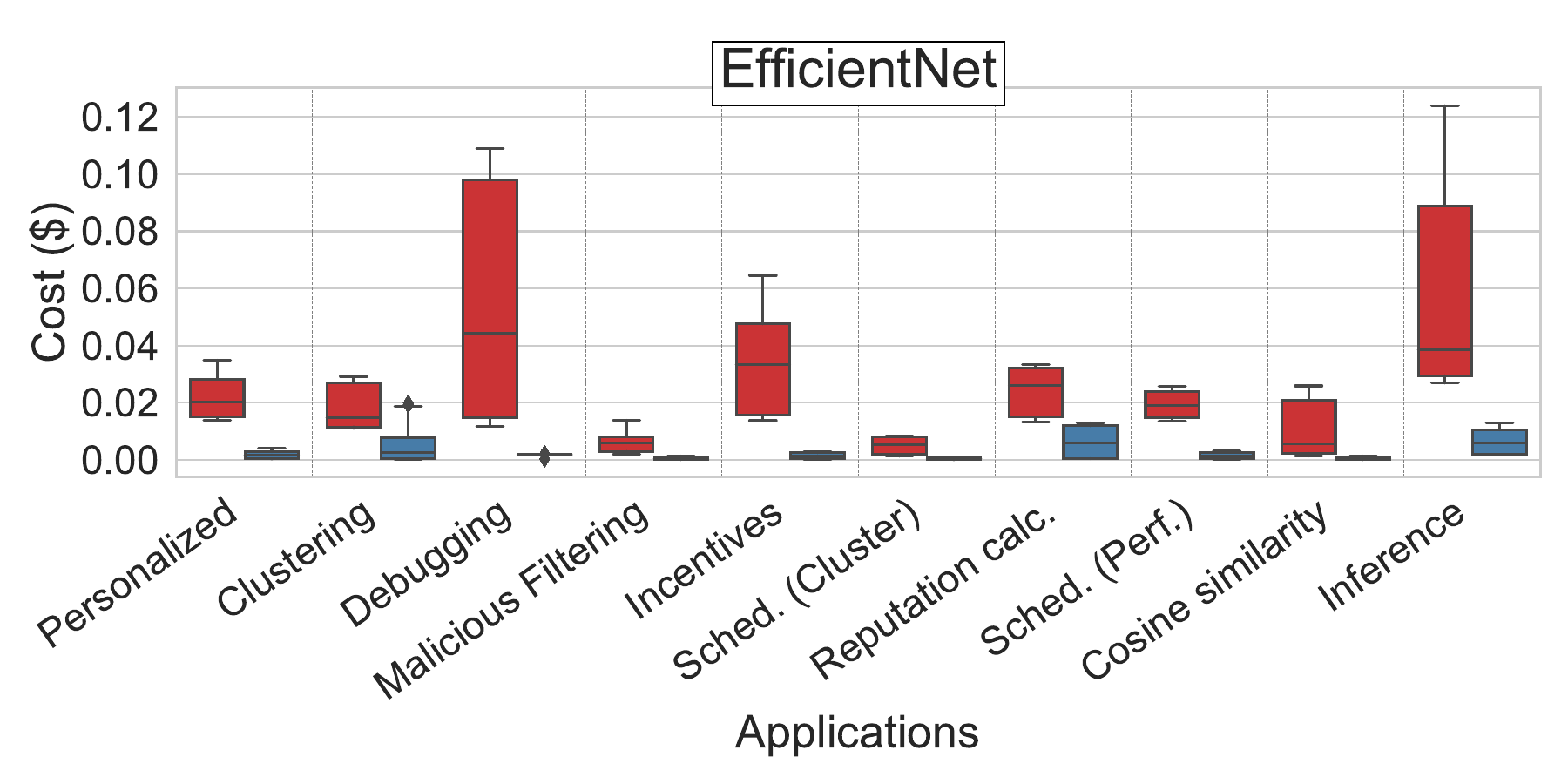}
   \vspace{-10pt}
   \label{subfig:FLStore_total_cost_EfficientNet}
\end{subfigure}
\begin{subfigure}[t]{0.49\textwidth}
  \centering
   \includegraphics[width=1\textwidth]{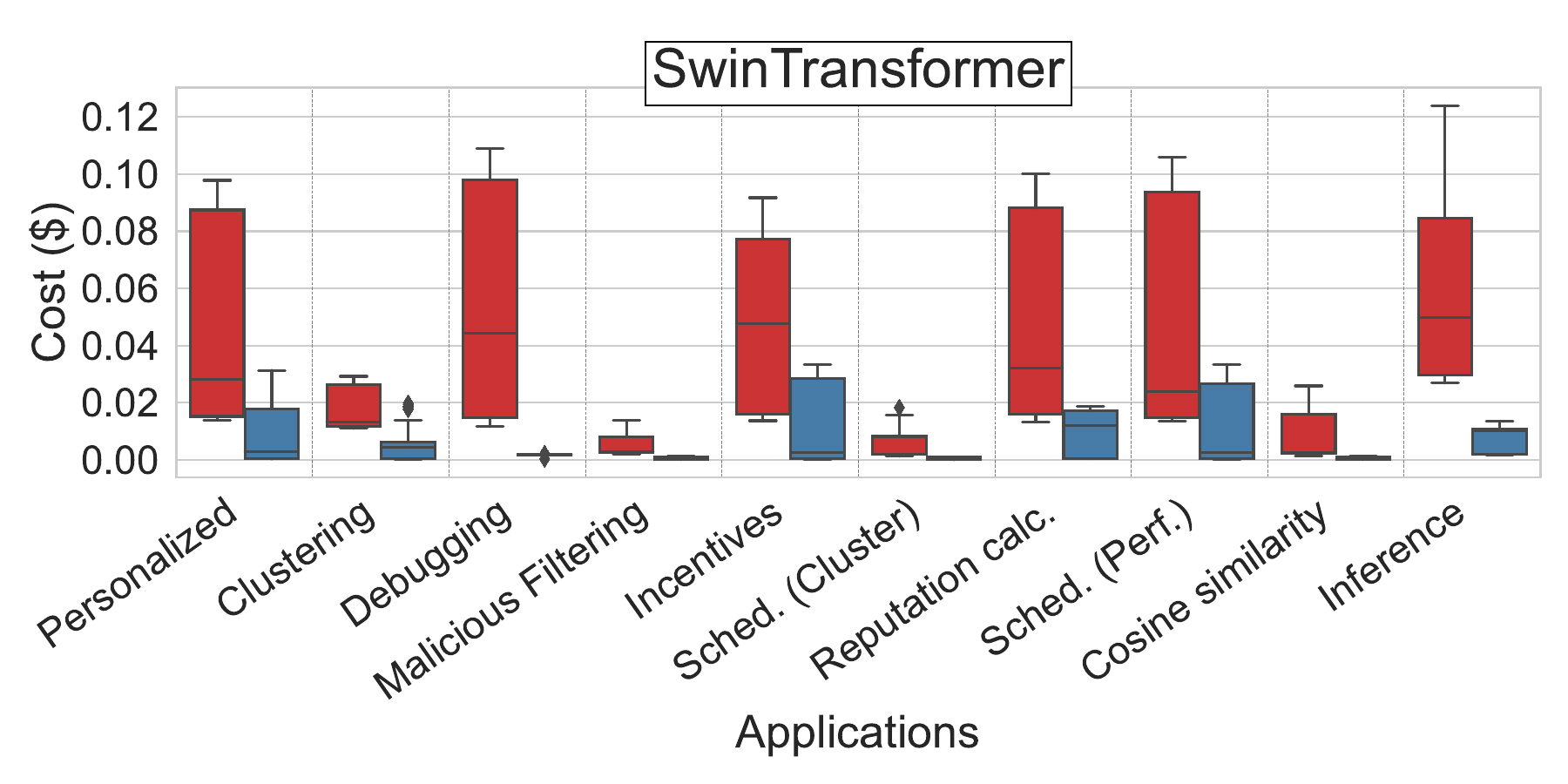}
   \vspace{-10pt}
   \label{subfig:FLStore_total_cost_SwinTransformer}
\end{subfigure}
  \caption{FLStore vs. ObjStore-Agg \textbf{per request cost} comparison over 50 hours.}
  \label{fig:per_request_cost_comparison}
\end{figure*}

  

\subsection{Latency Analysis}
\label{subsec: Latency Analysis}
\subsubsection{FLStore vs Cloud Object Store}
\label{subsubsec:FLStore vs Cloud Object Store (latency)}

We compare the latency and cost of baseline (ObjStore-Agg) and FLStore for ten workloads over 50 hours. Unlike ObjStore-Agg, FLStore co-locates the compute and data planes and utilizes tailored caching policies to cache relevant data in memory to reduce latency. However, in ObjStore-Agg the required data is fetched from the persistent store (data plane). Figure~\ref{fig:per_request_time_comparison} shows the latency for communication and computation per request. FLStore shows significant improvements in latency with its locality-aware computation and caching policies. On average, FLStore decreases the latency by $55.14$ seconds ($50.75\%$) per request, with up to $363.5$ seconds of maximum decrease ($99.94\%$) in latency per request. 
It can be observed in Figure~\ref{fig:per_request_time_comparison} that for some applications such as Incentives and Sched. (Perf.), SwinTransformer has a large distribution in the third quartile compared to ObjStore-Agg. However, FLStore still exhibits a lower median response time for these worklaods.

In distributed deep learning applications like FL, the main bottleneck is the increased communication time~\cite{MLSYS2019_communication_bottleneck,tang2023communicationefficientdistributeddeeplearning}. Thus, we analyze total latency (computation vs. communication) for the baseline (ObjStore-Agg) and our solution (FLStore) over 50 hours and 3000 non-training requests across 10 workloads. ObjStore-Agg is heavily communication-bound, with communication latency accounting for an average of $98.9\%$ of the total latency. 
FLStore mitigates this communication bottleneck improving the latency performance. With FLstore, we observe an average of $82.04\%$ ($35.50$ second) decrease in latency for Resnet18, $47.33\%$ ( $75.99$ second) for MobileNet, $50.44\%$ ($100.18$ second) for EfficientNet, and $20.45\%$ ($4.42$ second) decrease in latency for SwinTransformer compared to ObjStore-Agg. \textit{Due to space constraints, detailed results are provided in the Appendix.}

\subsubsection{FLStore vs In-Memory Cache}
\label{subsubsec: FLStore vs In-Memory Cache (latency)}

We also compare FLStore with other popular in-memory caching solutions available by cloud frameworks. Classic caching solutions like Redis and Memcached included in AWS ElastiCache allow for such in-memory caching~\cite{aws_elasticache}. Figure ~\ref{fig:elasticache_per_request_time_and_cost_comparison}, shows the result of the comparison between FLStore and AWS ElastiCache with AWS SageMaker baseline (Cache-Agg) per request.
It can be observed that per-request FLStore shows a $64.66\%$ on average and a maximum of $84.41\%$ reduction in latency when compared with Cache-Agg.  This reduction in latency is brought by co-located compute and data planes and locality-aware request processing in FLStore.

For the total latency breakup analysis over 50 hours and across 3000 non-training requests, FLStore shows a decrease in the total time by $37.77\%$ to $84.45\%$, amounting to a reduction of 191.65 accumulated hours for all requests. \textit{When comparing both Cache-Agg and ObjStore-Agg on the same workloads, FLStore shows an average decrease in latency of $71\%$ with ObjStore-Agg and $64.66\%$ with Cache-Agg. The larger reduction with ObjStore-Agg is due to cloud object stores being slower than cloud caches.}




\subsection{Cost Analysis}
\label{subsec:Cost Analysis}

\subsubsection{FLStore vs Cloud Object Store}
\label{subsubsec:FLStore vs Cloud Object Store (cost)}
In addition, we performed a per-request cost comparison across the ten selected workloads and 50 total hours. Figure~\ref{fig:per_request_cost_comparison} shows significant cost reduction with FLStore compared to ObjStore-Agg. The majority of this cost reduction stems from the reduced latency due to low data movement and the overall low computation cost of serverless functions for computation-light workloads. FLStore has an average cost decrease of $0.025$ cents per request with a maximum decrease of $0.094$ cents. On average, the cost of these applications in FLStore is  $88.23\%$ less than the cost of ObjStore-Agg baseline, with one application (Client Scheduling with Cosine Similarity for MobileNetV2) showing a $99.78\%$ decrease in per request cost.

We also performed the total cost breakup analysis over 50 hours, 3000 total non-training requests, and 10 workloads, calculating both the communication and computation costs for ObjStore-Agg and FLStore. We observe that the majority of the cost for ObjStore-Agg stems from the communication bottleneck. Resnet18, EfficientNet, SwinTransformer, and MobileNet spend $87.46\%$, $76.96\%$, $53.32\%$, and $85.80\%$ of their total latency respectively in communication. For the same settings, FLStore shows an average decrease of $94.73\%$, $92.72\%$, $77.83\%$, and $86.81\%$ in costs for Resnet18, MobileNet, SwinTransformer, and EfficientNet models respectively. Thus, FLStore significantly reduces the data transfer costs by unifying the compute and data planes. \textit{Due to
space constraints, Figures for these results are provided in the Appendix.}

\begin{figure}
\vspace{-5pt}
  \centering
{\includegraphics[width=0.5\columnwidth]{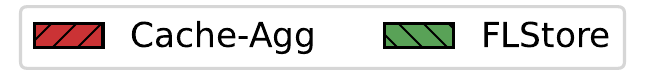}}
\vspace{-8.5pt}
\begin{subfigure}[t]{0.8\columnwidth}
  \centering
   \includegraphics[width=1\columnwidth]{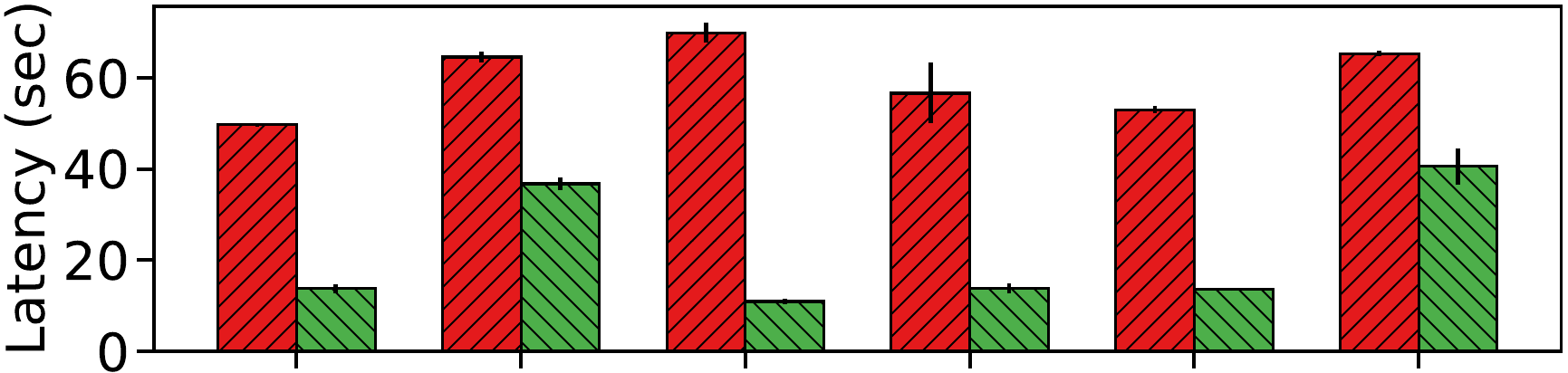}
   \label{subfig:ElastiCache_FLStore_time}
\end{subfigure}
\begin{subfigure}[t]{0.835\columnwidth}
  \centering
   \includegraphics[width=1\columnwidth]{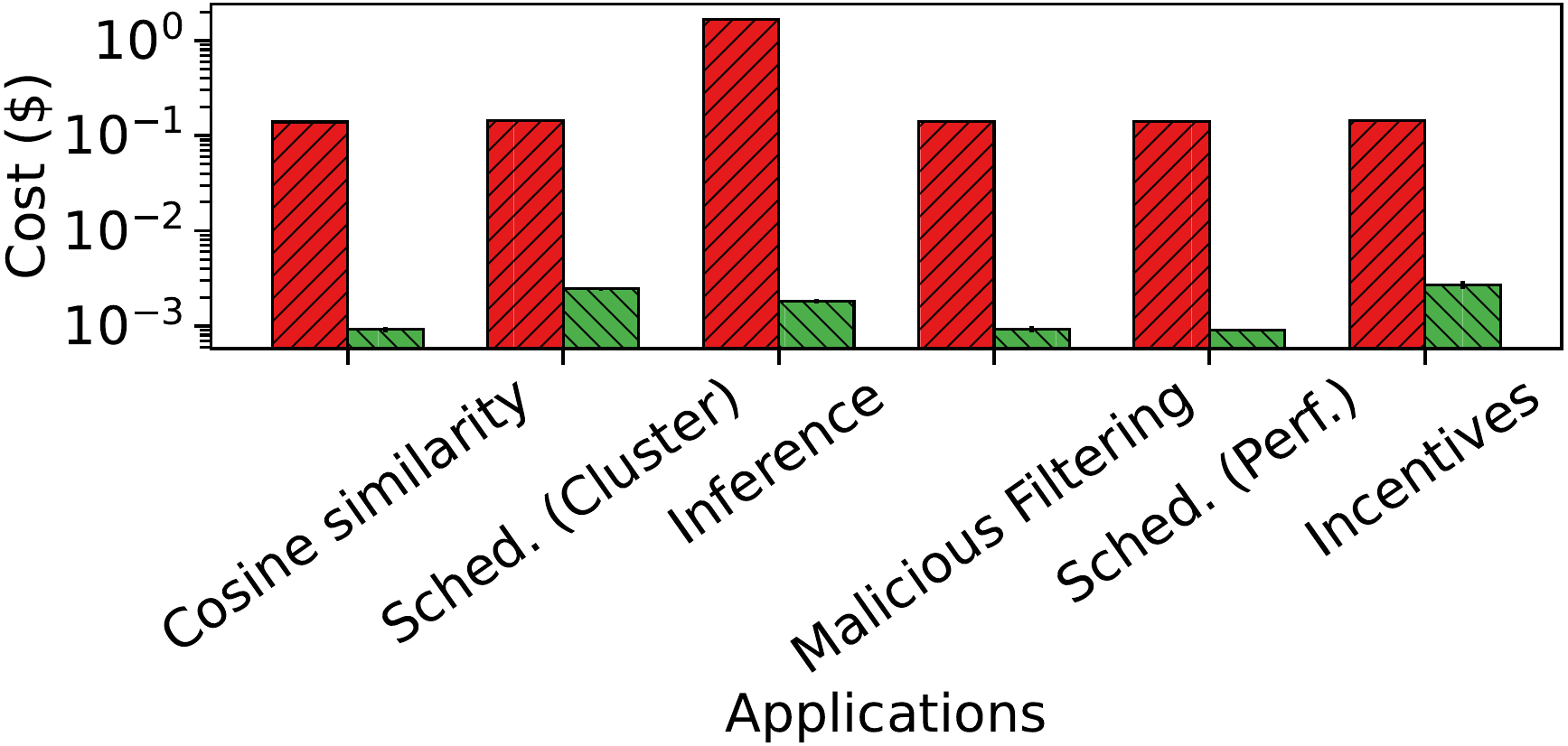}
   \vspace{-10pt}
   \label{subfig:ElastiCache_FLStore_cost}
\end{subfigure}
\hspace{5pt}
  \caption{Cache-Agg baseline vs. FLStore variants: \textbf{Per request latency} (top) and \textbf{cost} (bottom) over 50 hours.}
  \vspace{-4pt}
  \label{fig:elasticache_per_request_time_and_cost_comparison}
\end{figure}

\begin{figure}[t]
  \centering
  \includegraphics[width=\linewidth]{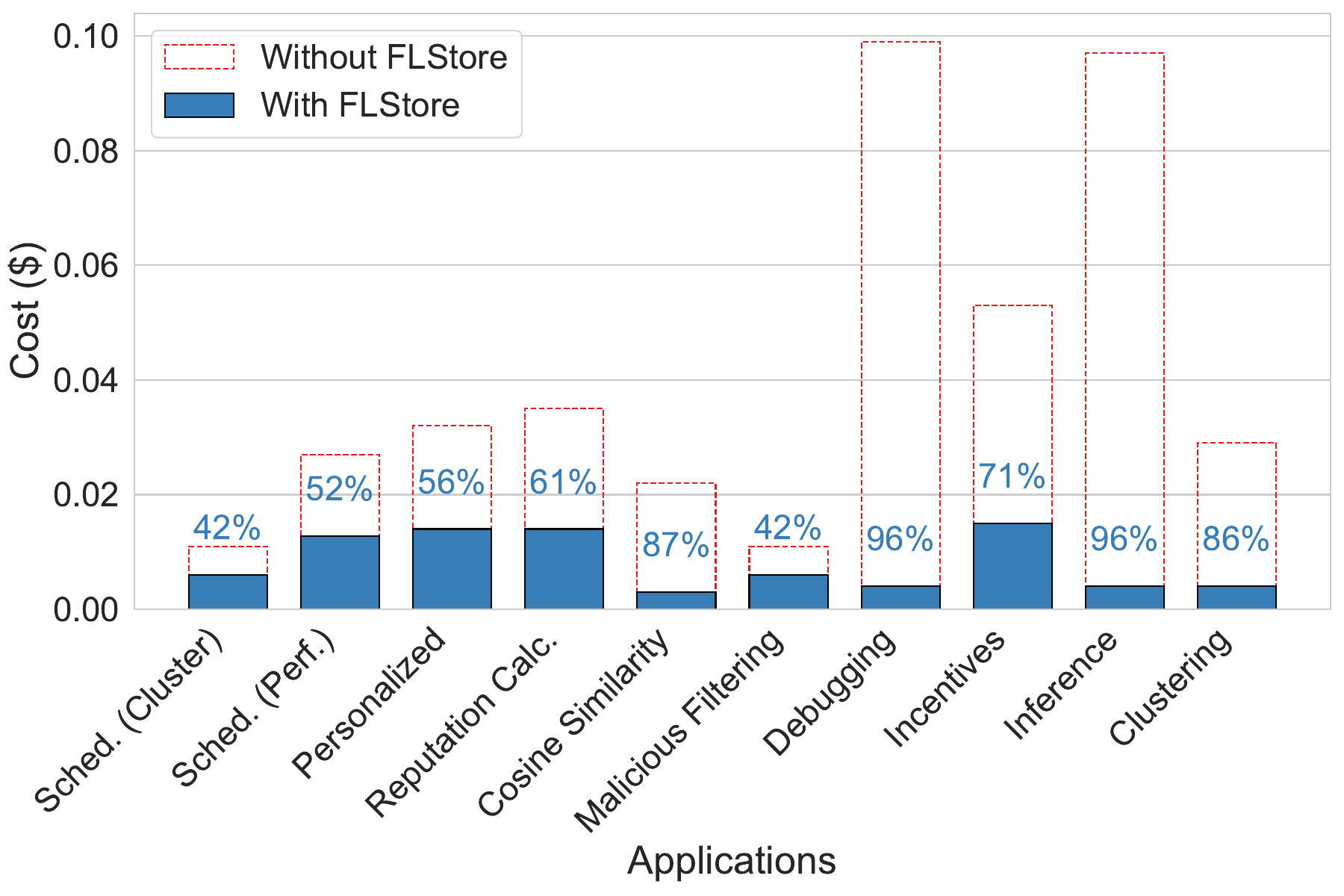}
  \vspace{-10pt}
  \caption{Overall cost of FL process per round with and without FLStore, with 200 clients, EfficientNet model~\cite{tan2021efficientnetv2}, 1000 training rounds, and CIFAR10 Dataset~\cite{krizhevsky2009learning}.}
  \label{fig:overall_vanilla_FL_vs_non_training_workloads_cost}
\end{figure}

\begin{figure*}

  \centering
{\includegraphics[width=0.7\textwidth]{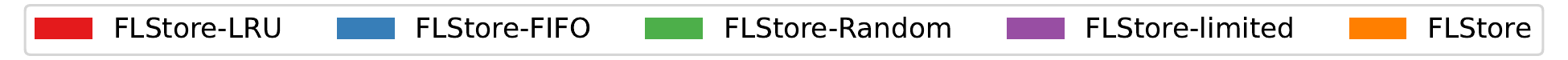}}
\vspace{-5pt}

\begin{subfigure}[t]{0.49\textwidth}
  \centering
   \includegraphics[width=1\textwidth]{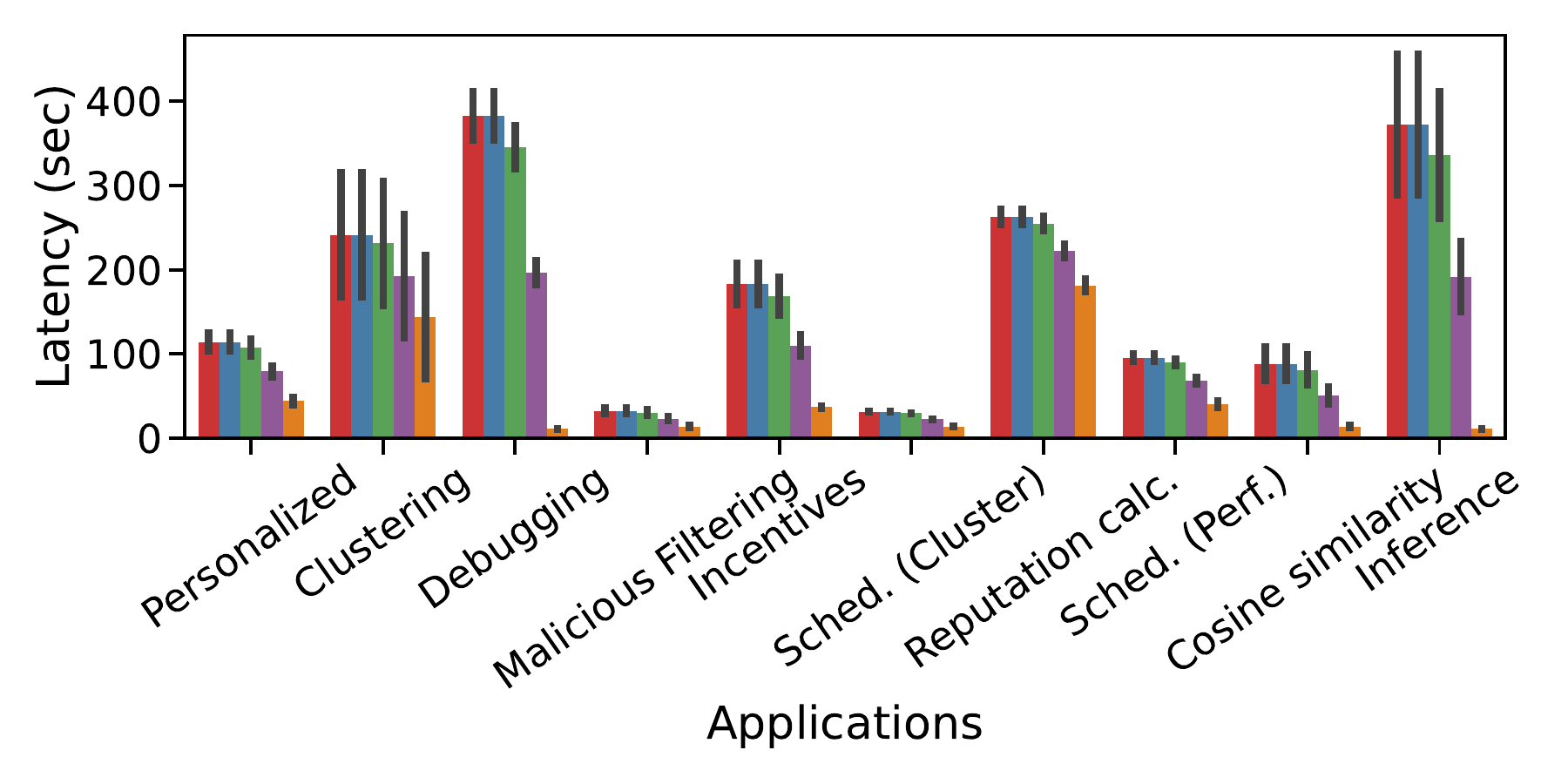}
   \vspace{-10pt}
   \label{subfig:other_caching_policies_FLStore_time_fault_tolerance}
\end{subfigure}
\begin{subfigure}[t]{0.49\textwidth}
  \centering
   \includegraphics[width=1\textwidth]{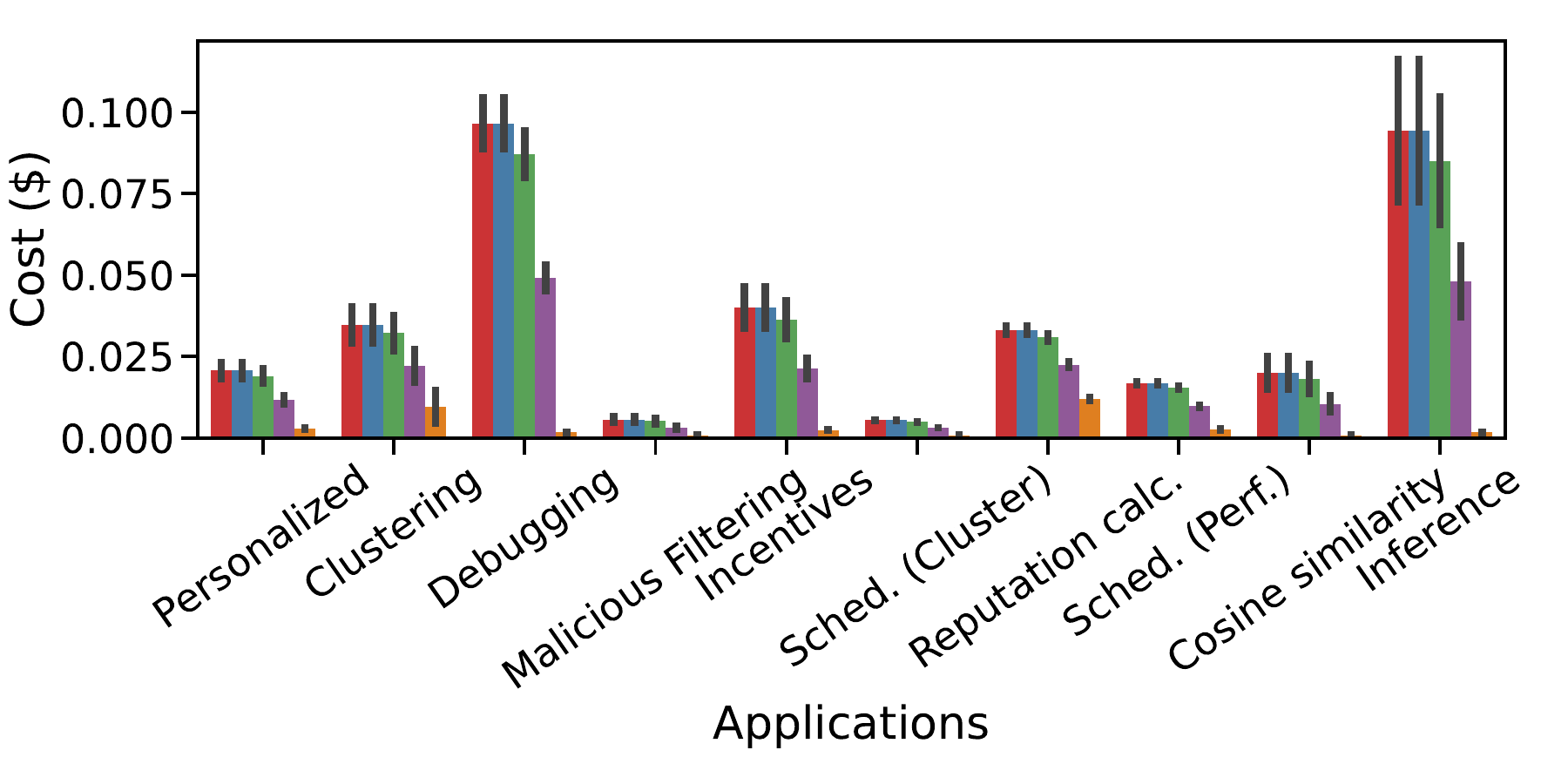}
   \vspace{-10pt}
   \label{subfig:other_caching_policies_FLStore_cost_fault_tolerance}
\end{subfigure}

\vspace{-12pt}
  \caption{\textbf{Per request latency} (left) and \textbf{cost} (right) comparison of various caching policies in FLStore over 50 hours.}
  \label{fig:other_caching_policies_total_cost_cost_comparison}
\end{figure*}


\subsubsection{FLStore vs In-Memory Cache}
\label{subsubsec: FLStore vs In-Memory Cache (cost)}

We can observe in Figure~\ref{fig:elasticache_per_request_time_and_cost_comparison} that keeping data in an in-memory cache such as ElastiCache is more costly in comparison to FLStore.
FLStore shows an average decrease of $98.83\%$ and a maximum decrease of $99.65\%$ in cost per request compared to Cache-Agg. This stems from the increased communication latency and costs because Cache-Agg does not have co-located computational resources for processing the cached data so the data still needs to be transferred to another cloud service such as AWS SageMaker~\cite{aws_sagemaker}.

For the cost breakup analysis over 50 hours and across 3000 non-training requests, FLStore shows a reduction of $98.12\%$ to $99.89\%$, resulting in accumulated savings of $\$7047.16$. Cloud caches tend to be more expensive than cloud object stores, which is why FLStore demonstrates an average cost decrease of $98.83\%$ when compared to Cache-Agg, and a $92.45\%$ decrease in cost when compared to ObjStore-Agg. \textit{The total time and total cost breakup analysis for both ObjStore-Agg and Cache-Agg is provided in the Appendix~\ref{sec:total performance breakup}.}

\subsubsection{Overall cost reduction with FLStore}
\label{subsubsec: Overall cost reduction with FLStore}


We also evaluated the overall reduction in FL costs brought by optimizing non-training workloads through FLStore.
Figure~\ref{fig:overall_vanilla_FL_vs_non_training_workloads_cost} shows that although FLStore does not directly reduce training costs, it significantly lowers the overall per-round cost by minimizing the communication overhead associated with non-training tasks. For example, debugging costs are reduced from $\$0.099$ to $\$0.004$ (a reduction of 96.4\%), and inference costs drop from $\$0.097$ to $\$0.004$ (a 96\% reduction). Other tasks, such as reputation calculation and cosine similarity, also see substantial cost savings.
These findings show that FLStore eliminates the need for frequent data transfers that typically inflate non-training costs also reducing the overall cost of FL jobs.

\begin{table}[t]
\vspace{-1.2em}
\centering
\caption{Cache Policy Performance Across Workloads}
\label{tab:cache_policy_performance}
\footnotesize
\begin{tabular}{p{0.33\columnwidth} p{0.1\columnwidth} p{0.06\columnwidth} p{0.06\columnwidth} p{0.06\columnwidth} p{0.06\columnwidth}}
\toprule
\textbf{Applications} & \textbf{Cache Policy} & \textbf{Hits} & \textbf{Misses} & \textbf{Total} & \textbf{Hit \%} \\
\midrule
\multirow{4}{*}{\begin{tabular}[t]{@{}l@{}} \cite{lai2021oort},\\ \cite{liu2022auxo}, \\ \cite{towards_personalized}, \\ \cite{shapleyFL} \end{tabular}} & FLStore (P2) & 19999 & 1 & 20000 & 0.99 \\
 & FIFO & 0 & 20000 & 20000 & 0 \\
 & LFU & 0 & 20000 & 20000 & 0 \\
 & LRU & 0 & 20000 & 20000 & 0 \\
\midrule
\multirow{4}{*}{\begin{tabular}[t]{@{}l@{}} \cite{fedDebug}, \\\cite{baracaldo2022accountable}, \\ \cite{tiff}, \\\cite{FedDNNDebugger} \end{tabular}} & FLStore (P3) & 63 & 1 & 64 & 0.98 \\
 & FIFO & 0 & 64 & 64 & 0 \\
 & LFU & 0 & 64 & 64 & 0 \\
 & LRU & 0 & 64 & 64 & 0 \\
\midrule
\multirow{4}{*}{\begin{tabular}[t]{@{}l@{}} \cite{Khan2024FLOAT}, \\\cite{NEURIPS2021_a0205b87}, \\ \cite{balta2021accountable}, \\\cite{lai2021oort} \end{tabular}} & FLStore (P4) & 20000 & 0 & 20000 & 1 \\
 & FIFO & 0 & 20000 & 20000 & 0 \\
 & LFU & 0 & 20000 & 20000 & 0 \\
 & LRU & 0 & 20000 & 20000 & 0 \\
\bottomrule
\end{tabular}
\vspace{-1em}
\end{table}

\subsection{FLStore vs Traditional Caching Policies}
\label{subsec:FLStore caching policies}

We introduce traditional caching strategies like Least Recently Used (LRU) and First In First Out (FIFO) in FLStore, alongside our tailored workload-specific policies derived from a developed taxonomy. We evaluate these against FLStore and its variant, FLStore-limited which depicts a limited storage availability scenario having half the storage capacity of FLStore. As depicted in Figure~\ref{fig:other_caching_policies_total_cost_cost_comparison}, both FLStore-LRU and FLStore-FIFO show similar performance due to their generic nature, unlike the taxonomy-driven policies of FLStore and FLStore-limited, which preemptively cache relevant data for imminent requests, thereby markedly reducing latency and costs. For instance, the debugging workload in Table~\ref{tab:combined_taxonomy_workloads} mandates the P2 caching policy, directing FLStore to cache the current training round's metadata rather than outdated information, leading to a significant reduction in debugging latency by $97.15\%$ (380 seconds) and cost savings of $\$0.1$ per request. Notably, even with limited capacity, FLStore-limited surpasses traditional policies. \textit{These improvements are substantial, especially given that the non-training requests can range from thousands to hundreds of thousands}.

We evaluated FLStore's performance against traditional caching policies like LFU, LRU, and FIFO using a simulated trace for non-training FL requests, crafted from FL jobs for 10 clients each round from a pool of 250 over 2000 rounds on popular FL frameworks like Oort~\cite{lai2021oort}, FedDebug~\cite{fedDebug}, REFL~\cite{refl}, and others~\cite{towards_personalized,baracaldo2022accountable,NEURIPS2021_a0205b87} that utilize non-training applications. As shown in Table~\ref{tab:cache_policy_performance}, FLStore's caching policy achieves a $99\%$ hit rate for Clustering~\cite{liu2022auxo} and Personalized FL~\cite{towards_personalized} under the P2 caching policy and $98\%$ hit rate for tasks under the P3 caching policy~\cite{fedDebug,FedDNNDebugger, baracaldo2022accountable} with similar results observed for the P4 policy workloads~\cite{Khan2024FLOAT,NEURIPS2021_a0205b87,balta2021accountable}. In contrast, traditional policies consistently register a $0\%$ hit rate across all tested scenarios.

\textbf{Ablation study.} We also evaluated FLStore variants without tailored caching policies: FLStore-Random and FLStore-Static. FLStore-Random, using random caching policy selection regardless of workload, shows lower latency in some cases, as depicted in Figure~\ref{fig:other_caching_policies_total_cost_cost_comparison}. However, for critical workloads like Scheduling and Incentivization, its performance aligns with FLStore-FIFO and FLStore-LRU. Comparison with FLStore-Static is detailed in Appendix~\ref{sec:FLStore static: Ablation Study}.

\subsection{Overhead of FLStore's components}
\label{subsec: Cache Engine and Request Tracker Overhead}

The Cache Engine and Request Tracker can run co-located with the aggregator service or locally, with minimal overhead. We measure the overhead for $1000$ concurrent non-training requests. The Request Tracker uses less than $0.19$ MB of memory, and the Cache Engine uses $0.6$ MB. Scaling to $100000$ requests increases memory usage to $20.3$ MB and $63.2$ MB, respectively. In both cases, the time to retrieve, use, or remove data from these services is \textbf{under one millisecond}. The minimal overhead of the Cache Engine and Request Tracker allows them to be run locally, on the aggregator server, or even on a serverless function.

\section{Conclusion}
\label{sec:Conclusion}

This paper introduces FLStore, an efficient and cost-effective storage solution with locality-aware processing for FL's communication-heavy non-training workloads. 
Our experiments demonstrate that FLStore is efficient and cost-effective compared to other caching and cloud storage solutions. FLStore is scalable and robust and can incorporate new workloads by adding a new caching policy.
\vspace{-1em}
\section*{Acknowledgements}
This work is supported in part by the Amazon ML Systems Fellowship, University of Minnesota Grants to Advance Graduate Education (GAGE) Fellowship, Samsung Global Research Outreach (GRO) Award, NSF grants CSR-2106634 and CSR-2312785, and UK Research and Innovation (UKRI) - Engineering and Physical Science Research Council (EPSRC) under grant No. EP/X035085/1.




\bibliography{bibs/ali,bibs/FLStore,bibs/proposal}
\bibliographystyle{mlsys2025}


\appendix
\begin{figure*}[!ht]
  \centering
\includegraphics[width=0.8\textwidth]{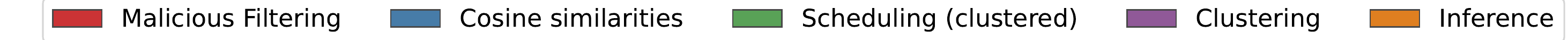}
\vspace{-2pt}

\begin{subfigure}[t]{0.49\textwidth}
  \centering
   \includegraphics[width=1\textwidth]{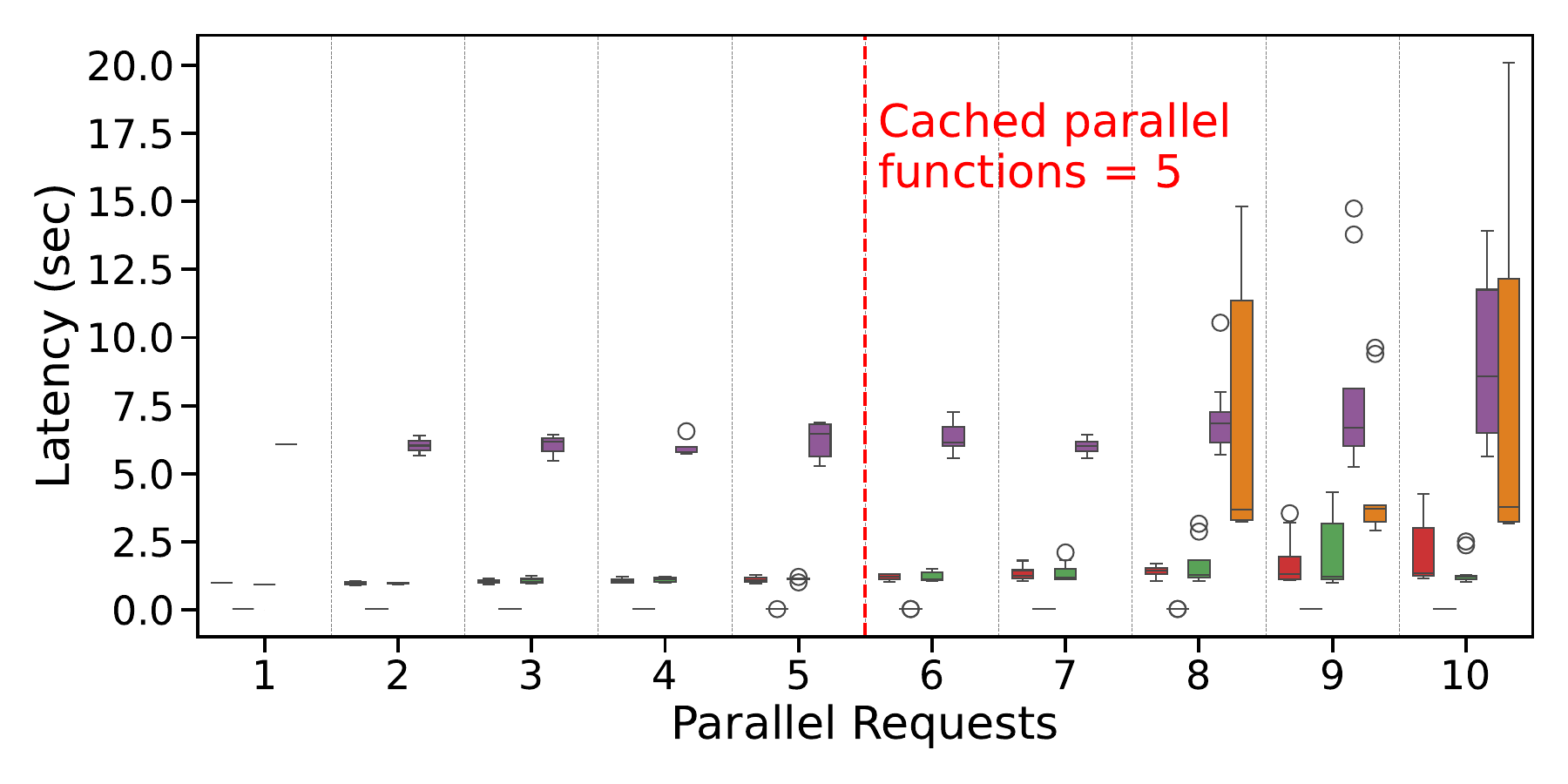}
   \vspace{-10pt}
   \label{subfig:FLStore_time_scalability}
\end{subfigure}
\begin{subfigure}[t]{0.49\textwidth}
  \centering
   \includegraphics[width=1\textwidth]{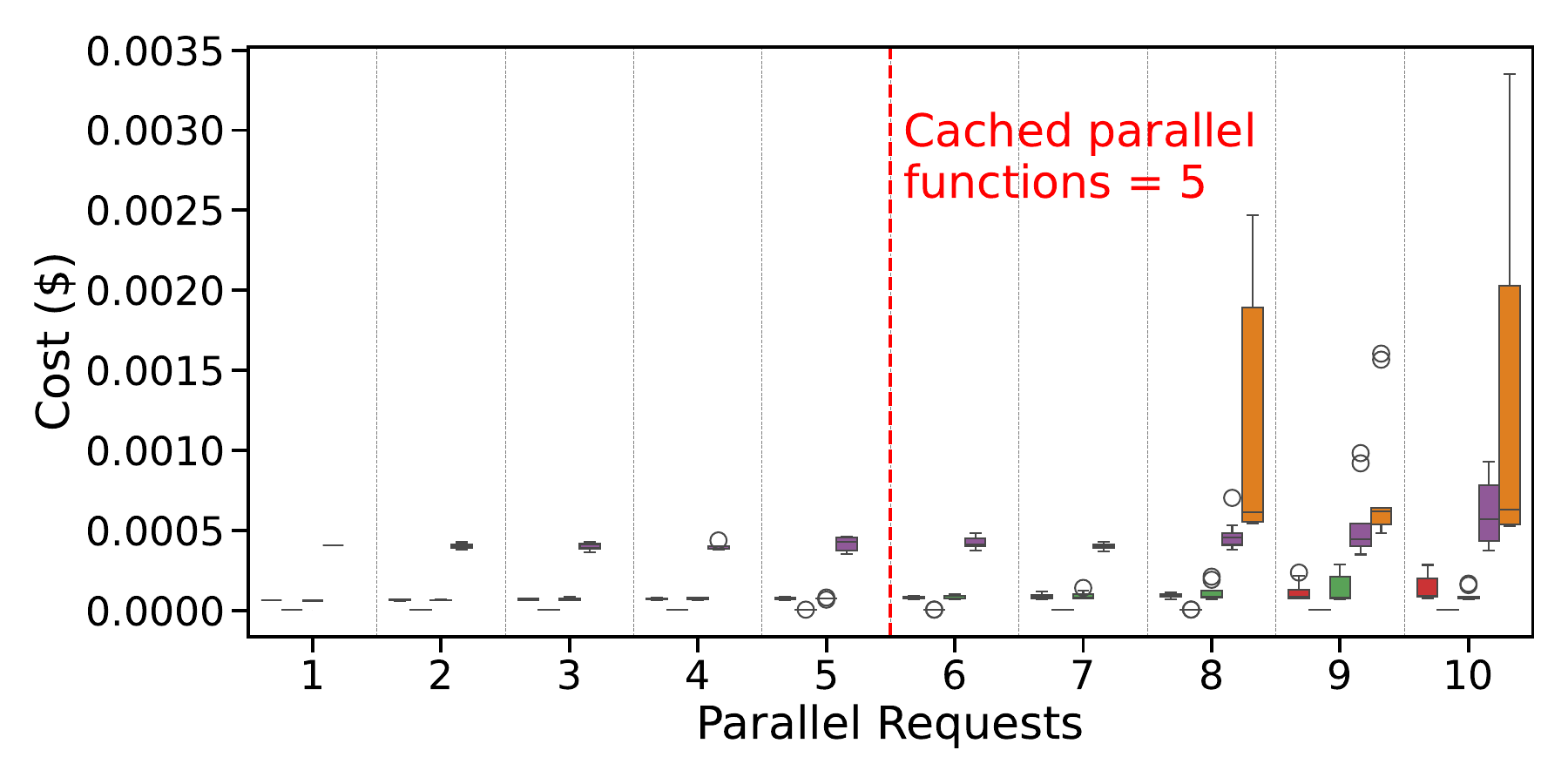}
   \vspace{-10pt}
   \label{subfig:FLStore_cost_scalability}
\end{subfigure}
  \caption{FLStore \textbf{scalability} for iteratively increasing parallel requests and 5 parallel cached functions.}
  \label{fig:FLStore_scalability}
\end{figure*}
\begin{figure*}[h]
  \centering
{\includegraphics[width=0.45\textwidth]{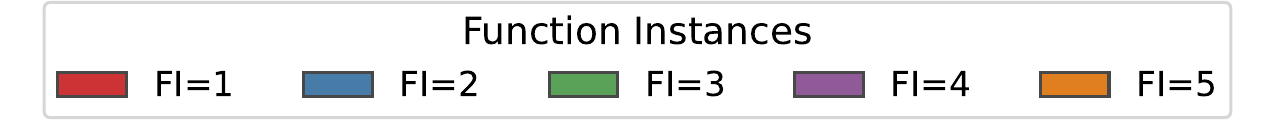}}
\vspace{-5pt}

\begin{subfigure}[t]{0.49\textwidth}
  \centering
   \includegraphics[width=1\textwidth]{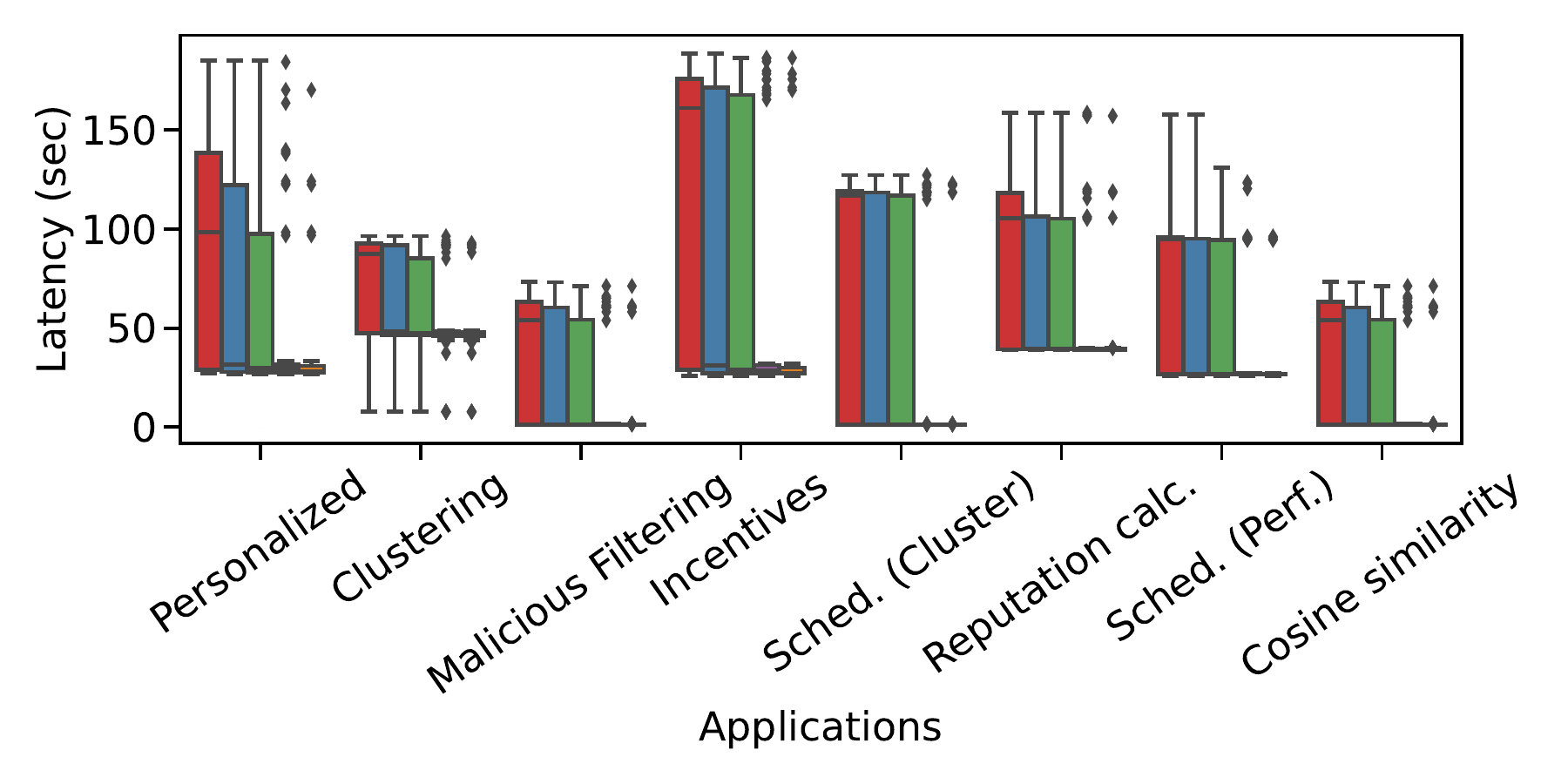}
   \vspace{-10pt}
   \label{subfig:FLStore_time_fault_tolerance}
\end{subfigure}
\begin{subfigure}[t]{0.49\textwidth}
  \centering
   \includegraphics[width=1\textwidth]{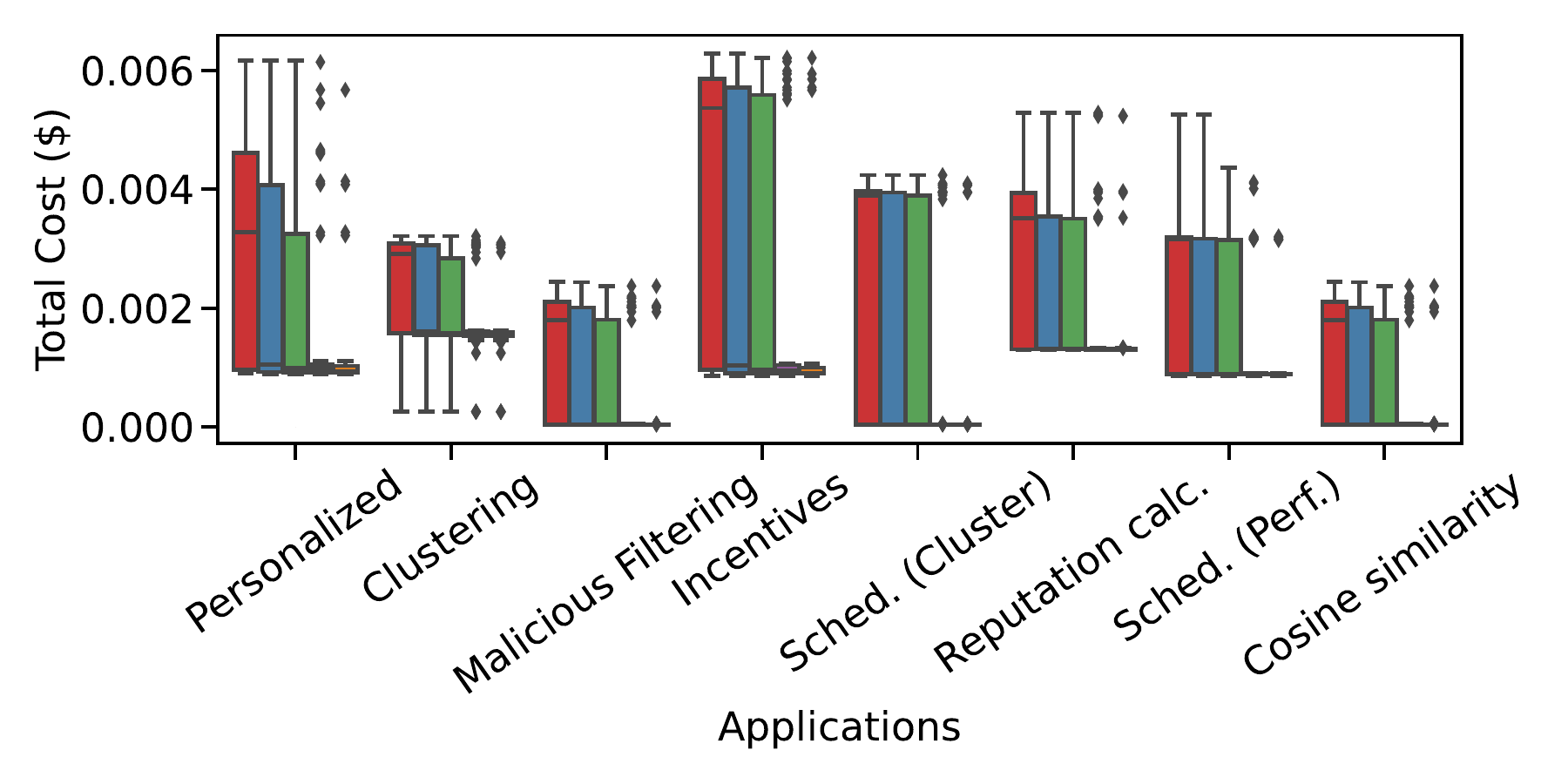}
   \vspace{-10pt}
   \label{subfig:FLStore_cost_fault_tolerance}
\end{subfigure}
  \caption{FLStore latency and cost per request over 50 hours with varying function instances (FI) for \textbf{fault tolerance}.}
  \label{fig:total_time_cost_fault_tolerance_comparison}
\end{figure*}
\newpage

\section{Supplementary Features of FLStore}
\label{sec:Supplementary Features}

\paragraph{Modular design} FLStore's modular design enables seamless integration with existing FL frameworks without modifying clients or aggregators. Training can proceed unchanged, while client updates and metadata received by the aggregator are asynchronously relayed to FLStore’s cache. FLStore then serves as a scalable and efficient storage solution, handling non-training tasks.

\paragraph{Multi-tenancy}
The serverless computing paradigm inherently provides isolation~\cite{aws_lambda,openfaas_faas}, allowing each user to create an isolated cache on the same FLStore instance. This enables customized caching policies per non-training workload/application, allowing FLStore to handle requests from multiple users simultaneously.

\begin{figure*}
\begin{subfigure}[t]{0.33\textwidth}
  \centering
   \includegraphics[width=1\textwidth]{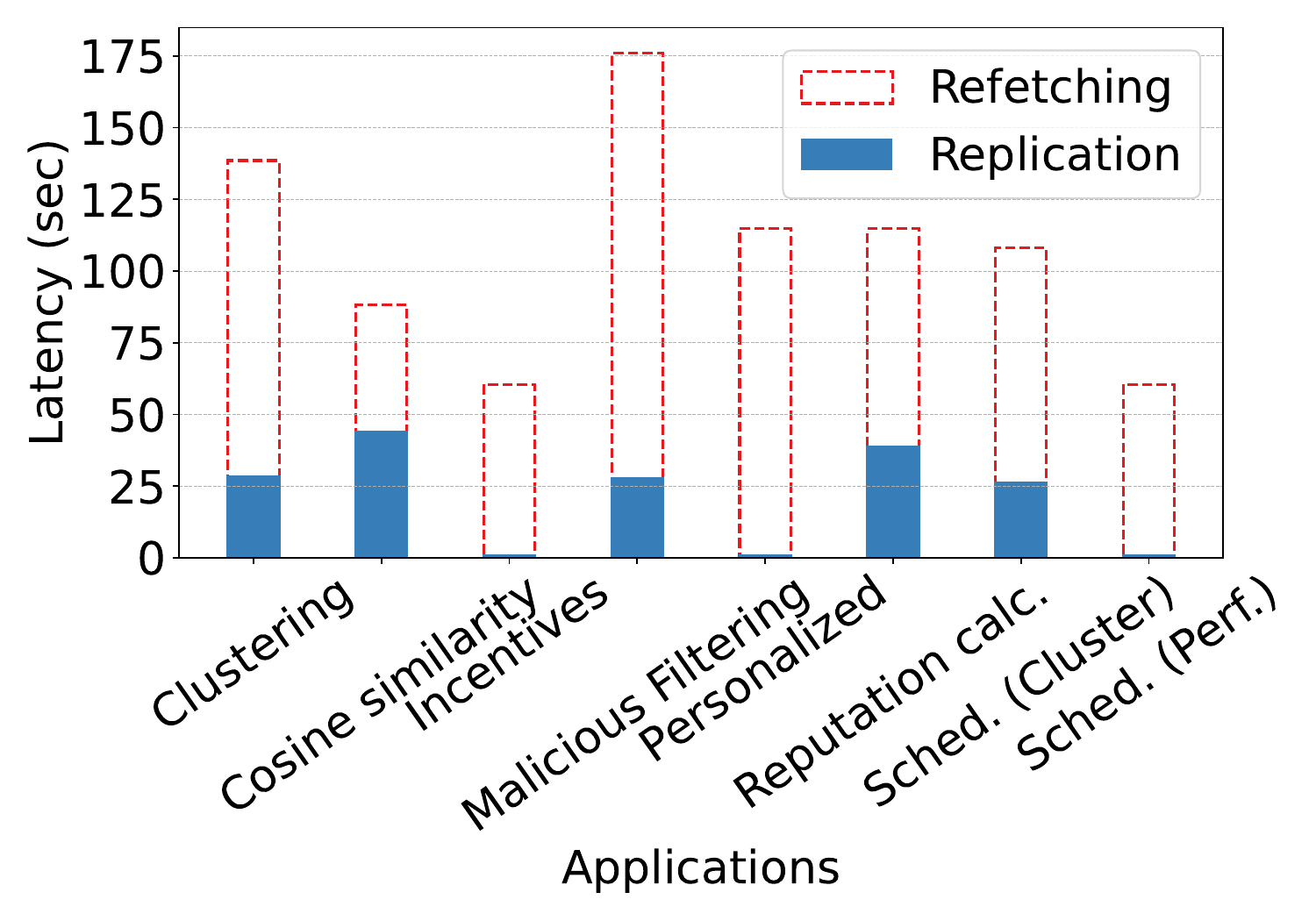}
   \vspace{-10pt}
   \label{subfig:replication_vs_recomputation_time}
\end{subfigure}
\begin{subfigure}[t]{0.33\textwidth}
  \centering
   \includegraphics[width=1\textwidth]{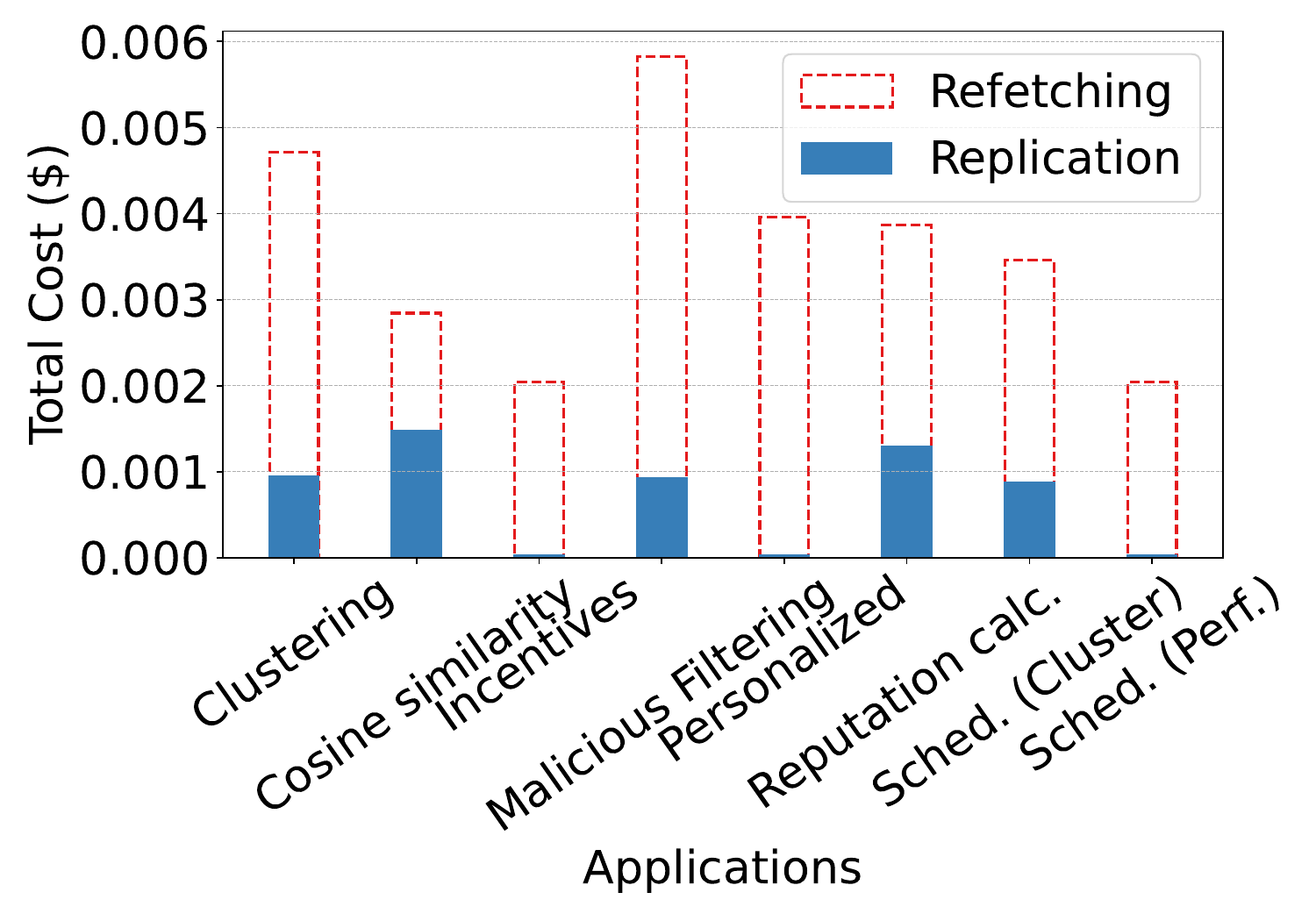}
   \vspace{-10pt}
   \label{subfig:replication_vs_recomputation_cost}
\end{subfigure}
\begin{subfigure}[t]{0.33\textwidth}
  \centering
   \includegraphics[width=1\textwidth]{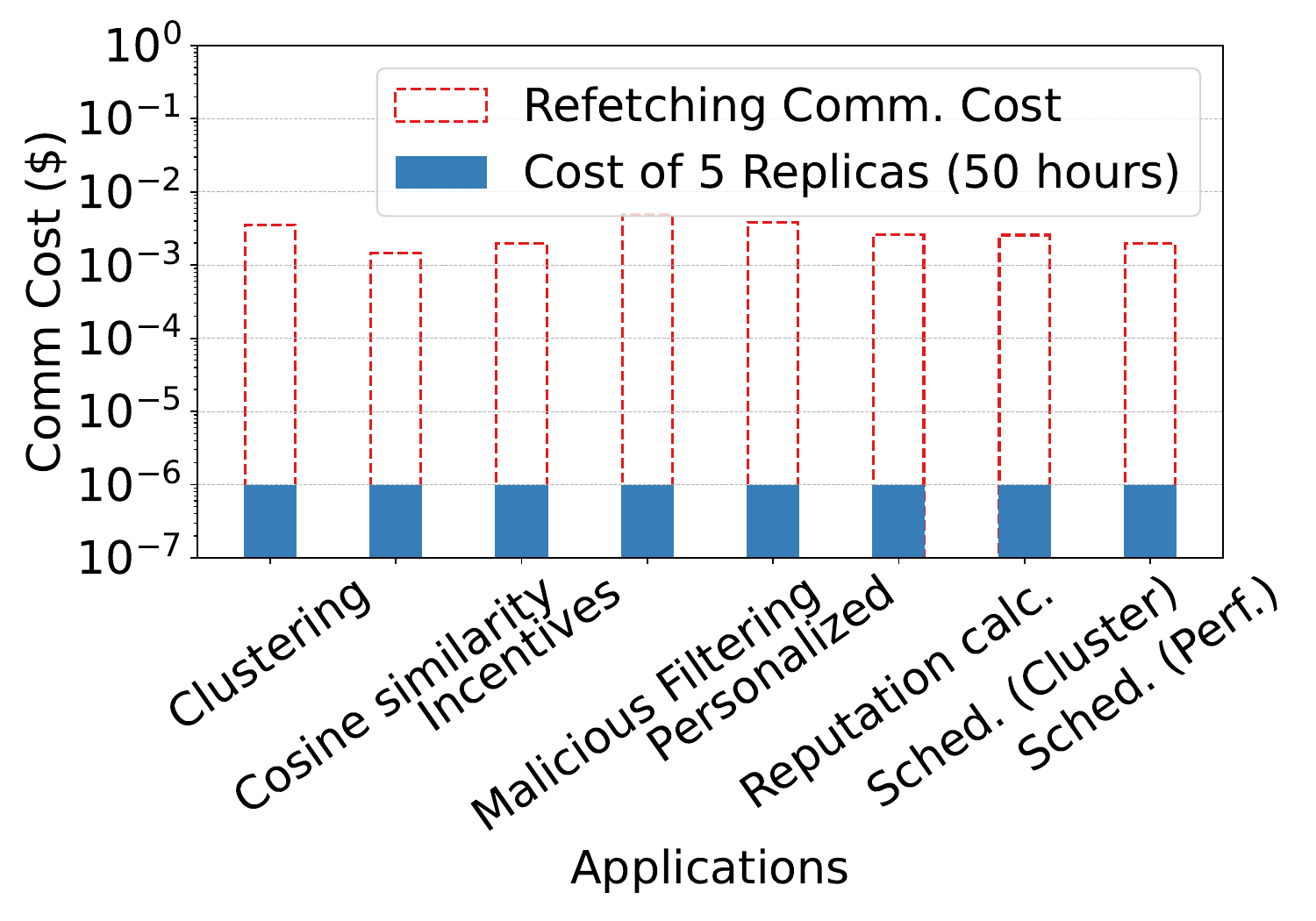}
   \vspace{-10pt}
   \label{subfig:replication_vs_recomputation_comm_cost}
\end{subfigure}
  \caption{Overall latency and cost comparison of replication vs. re-fetching (first and second from left), and communication cost comparison (rightmost).}
  \label{fig:replication_vs_recomputation}
\end{figure*}

\subsection{Scalability of FLStore}
\label{subsec:Scalability of FLStore}

To demonstrate FLStore's scalability, we simulated increasing concurrent non-training requests, with FLStore maintaining 5 cached function instances (red line, Figure~\ref{fig:FLStore_scalability}). We varied the number of concurrent client requests from 1 to 10 across six representative non-training workloads using the EfficientNet model.
As shown in Figure~\ref{fig:FLStore_scalability}, latency and cost remain nearly constant when concurrent requests are equal to or fewer than the cached functions. For 1 to 5 requests, the average latencies were $1.05$ seconds for Malicious Filtering, $0.031$ seconds for Cosine Similarities, $1.039$ seconds for Scheduling (clustered), and $6.067$ seconds for Clustering. Even with $6$ and $7$ requests, there was minimal increase in latency or cost. For $8$ to $10$ requests, latencies start increasing. However, this can be easily mitigated by scaling cached functions (creating copies of already cached functions) linearly with the number of requests, which incurs minimal additional cost, as discussed next.

\vspace{-3pt}
\subsection{Fault Tolerance}
\label{subsec:Fault Tolerance}
We evaluated FLStore's fault tolerance by testing ten different workloads using the EfficientNet model and sending $3000$ requests over $50$ hours. Faults (function reclamations) were generated based on the Zipfian distribution, observed in measurement studies on AWS Lambda~\cite{infinicache}. Figure~\ref{fig:total_time_cost_fault_tolerance_comparison} shows that with only $1$ function instance, latency and cost are highest, with improvement as the number of replicas increases. With $3$ to $5$ function instances, latency and cost remain nearly constant, despite faults. In particular, $3$ instances reduce latency by 50-150 seconds per request compared to a single instance in the face of faults.

Interestingly the cost of maintaining function replicas is negligible compared to the overhead and cost of re-computation and communication due to faults. For $50$ hours and $3000$ requests, maintaining $5$ replicas costs just $\$0.003$, or $\$0.000001$ per non-training request served (Figure~\ref{fig:replication_vs_recomputation}). In contrast, fewer instances lead to higher overhead and costs while maintaining more replicas reduces these costs by up to $3000\times$. Notably, we did not evaluate the impact of regular pinging, as this has already been explored in prior
works~\cite{zhang2023infinistore,infinicache}.

\begin{figure*}[t]
  \centering
\includegraphics[width=0.6\textwidth]{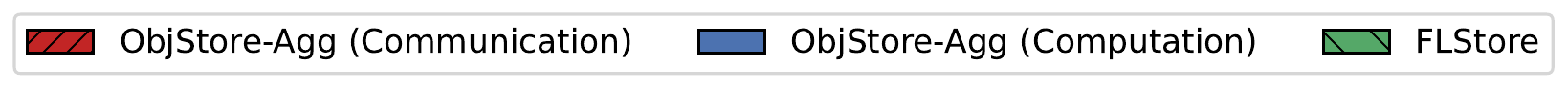}

\begin{subfigure}[t]{0.485\textwidth}
  \centering
   \includegraphics[width=1\textwidth]{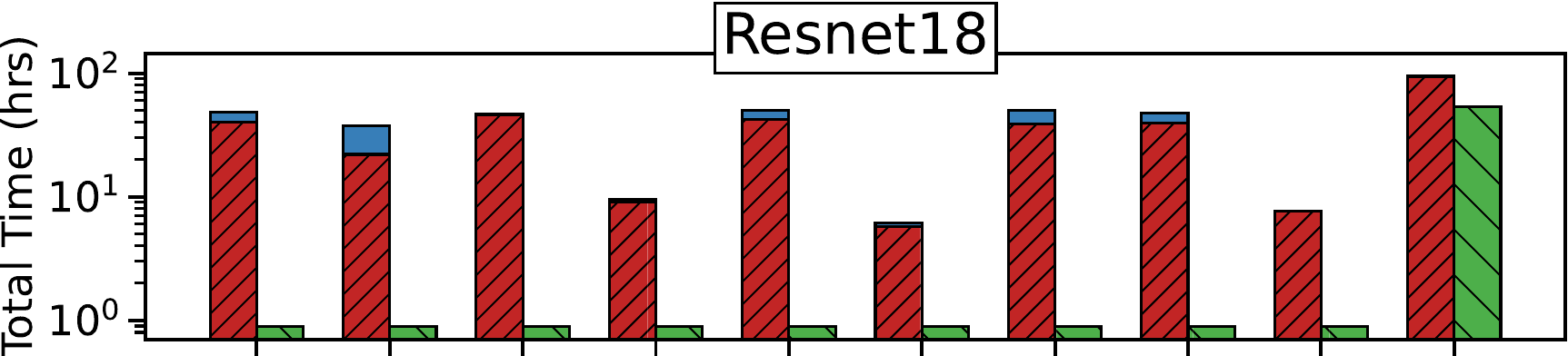}
   \label{subfig:FLStore_total_time_breakup_resnet18}
\end{subfigure}
\begin{subfigure}[t]{0.49\textwidth}
  \centering
   \includegraphics[width=1\textwidth]{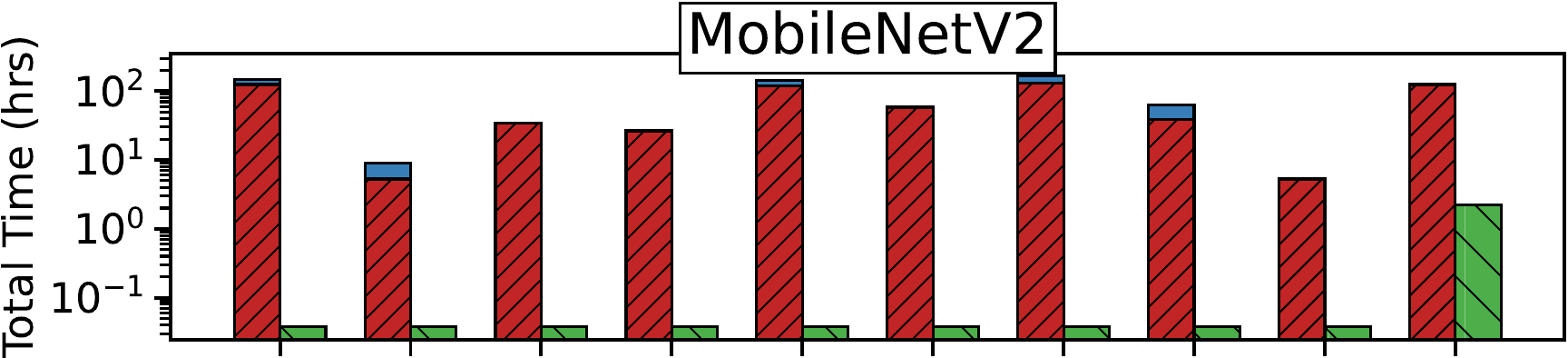}
   \label{subfig:FLStore_total_time_breakup_MobileNetV2}
\end{subfigure}
\begin{subfigure}[t]{0.49\textwidth}
  \centering
   \includegraphics[width=1\textwidth]{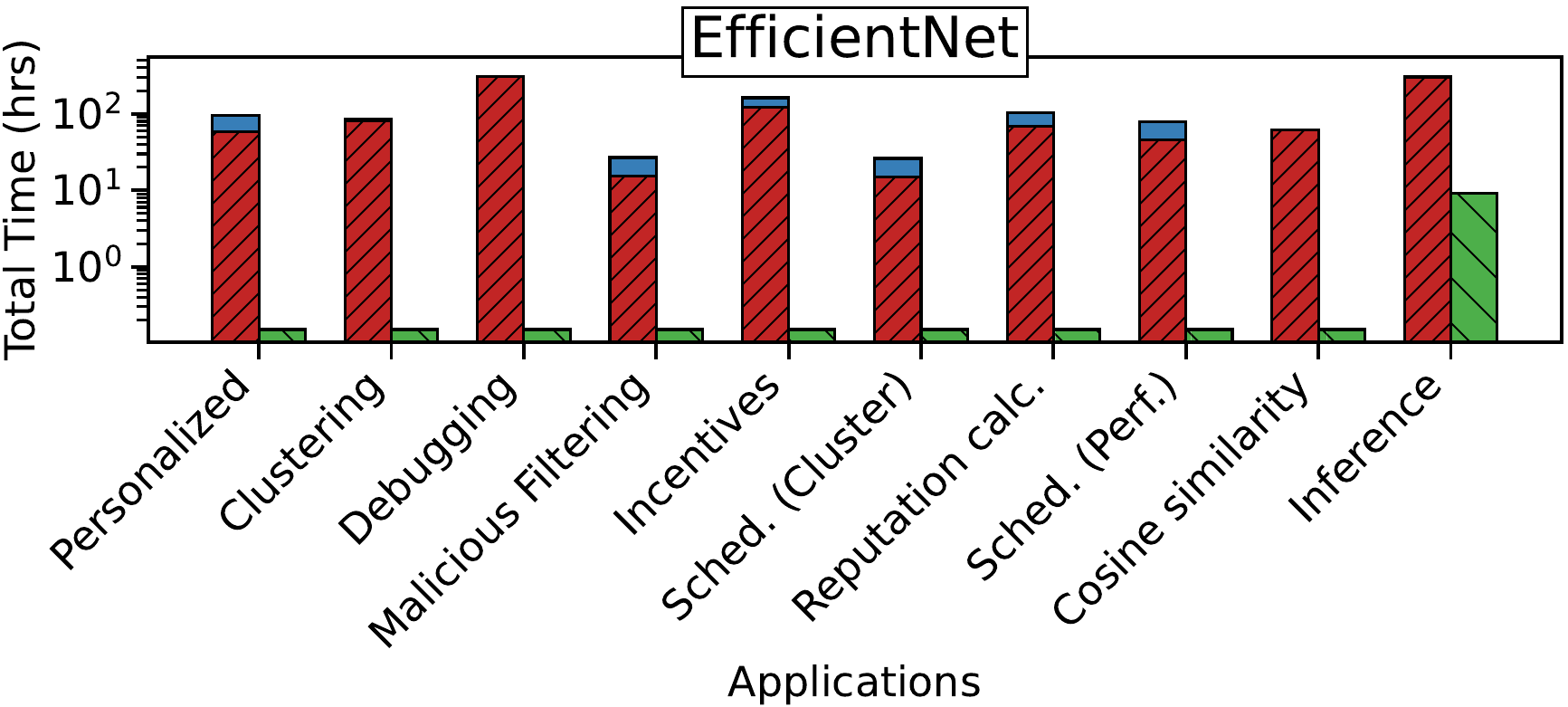}
   \label{subfig:FLStore_total_time_breakup_EfficientNet}
\end{subfigure}
\begin{subfigure}[t]{0.49\textwidth}
  \centering
   \includegraphics[width=1\textwidth]{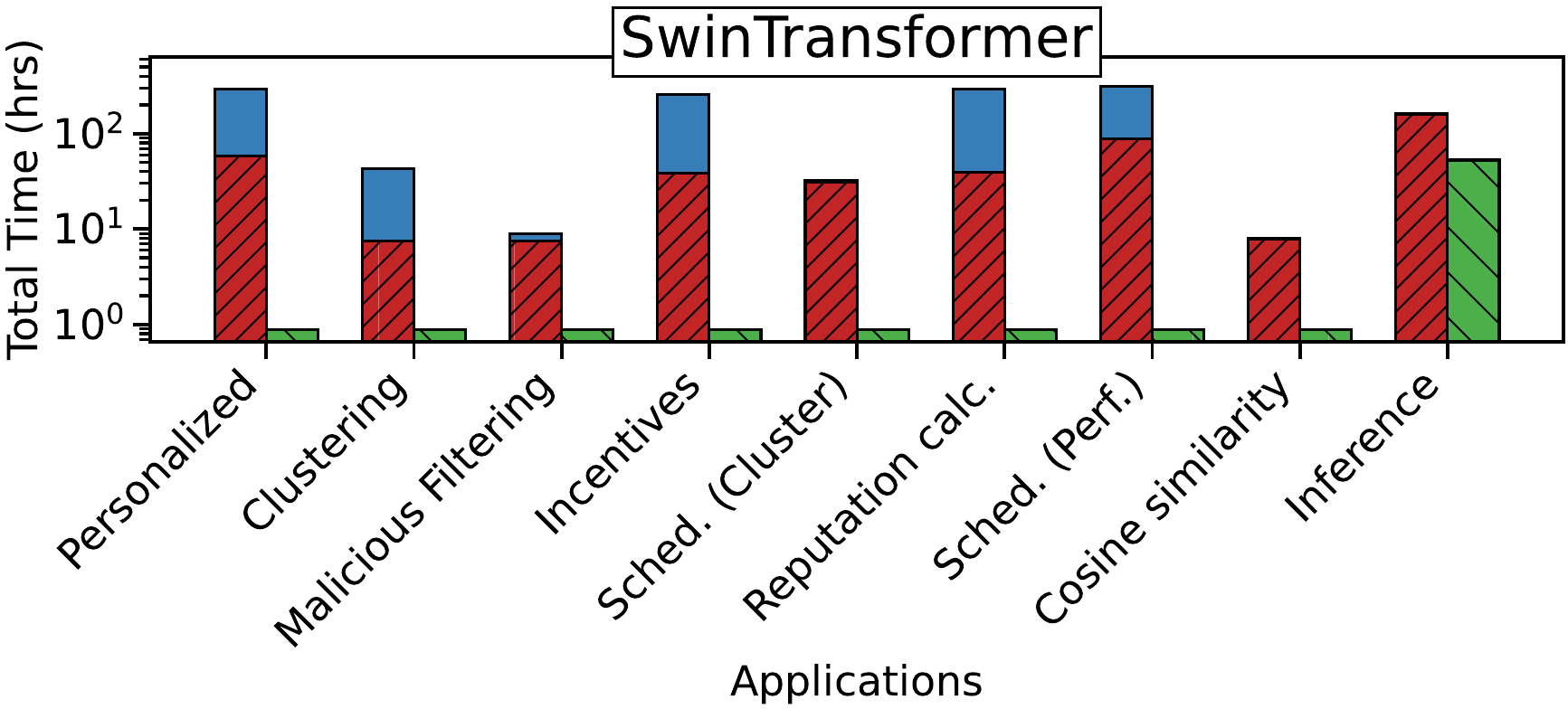}
   \label{subfig:FLStore_total_time_breakup_SwinTransformer}
\end{subfigure}
  \caption{FLStore vs. ObjStore-Agg \textbf{total time} breakup comparison over 50 hours.}
  
  \label{fig:total_time_breakup_comparison}
\end{figure*}
\begin{figure*}[!h]
  \centering
\includegraphics[width=0.6\textwidth]{figures/legend_breakup.pdf}

\begin{subfigure}[t]{0.48\textwidth}
  \centering
   \includegraphics[width=1\textwidth]{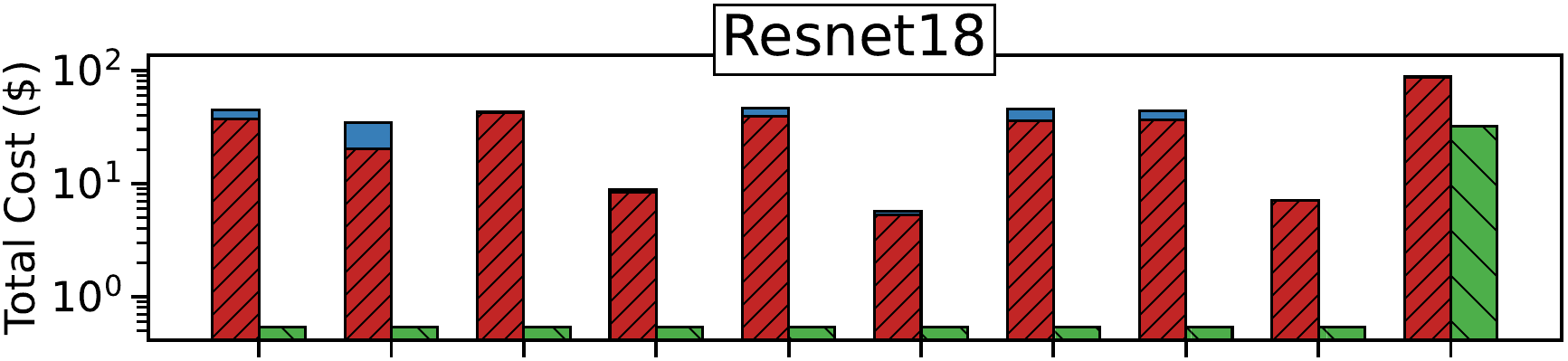}
   \label{subfig:FLStore_total_cost_breakup_resnet18}
\end{subfigure}
\begin{subfigure}[t]{0.49\textwidth}
  \centering
   \includegraphics[width=1\textwidth]{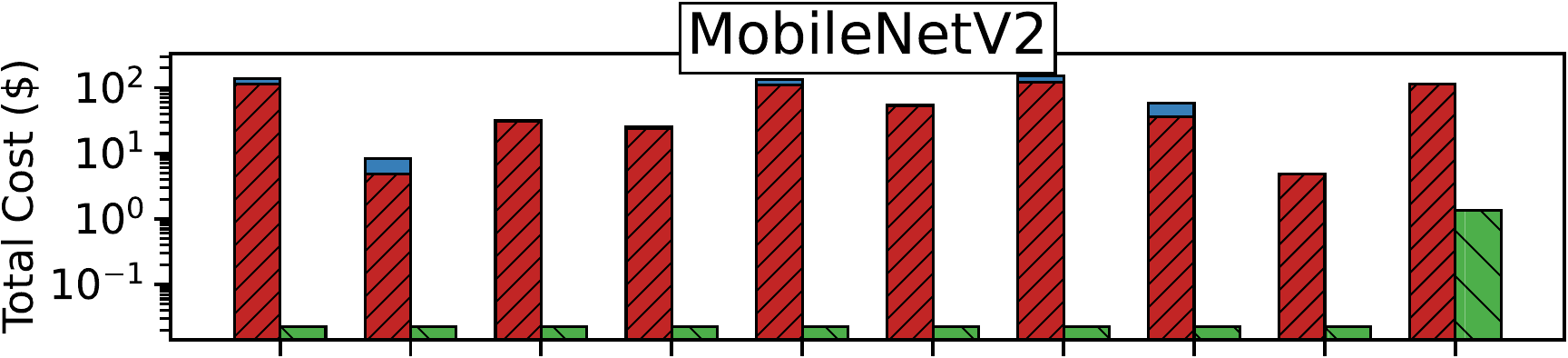}
   \label{subfig:FLStore_total_cost_breakup_MobileNetV2}
\end{subfigure}
\begin{subfigure}[t]{0.49\textwidth}
  \centering
   \includegraphics[width=1\textwidth]{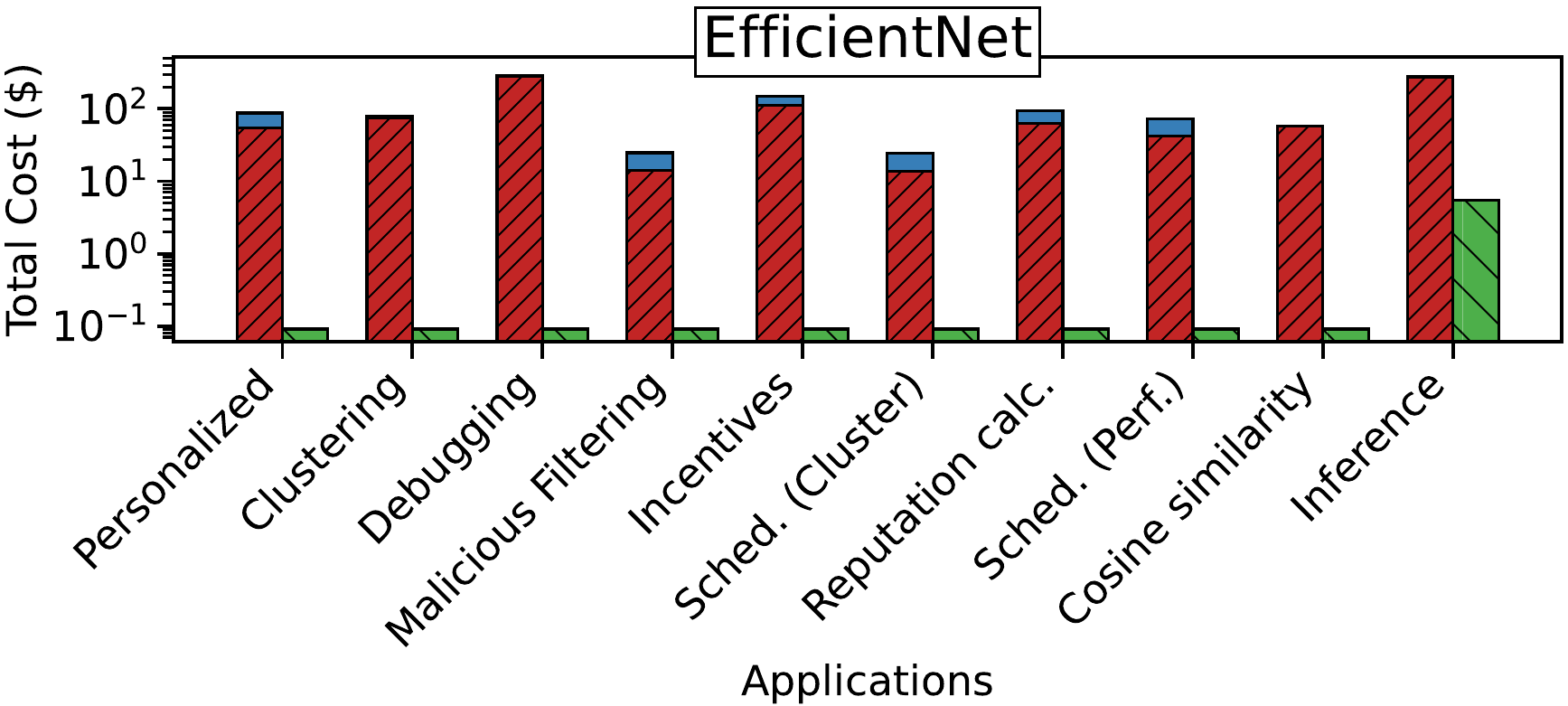}
   \label{subfig:FLStore_total_cost_breakup_EfficientNet}
\end{subfigure}
\begin{subfigure}[t]{0.49\textwidth}
  \centering
   \includegraphics[width=1\textwidth]{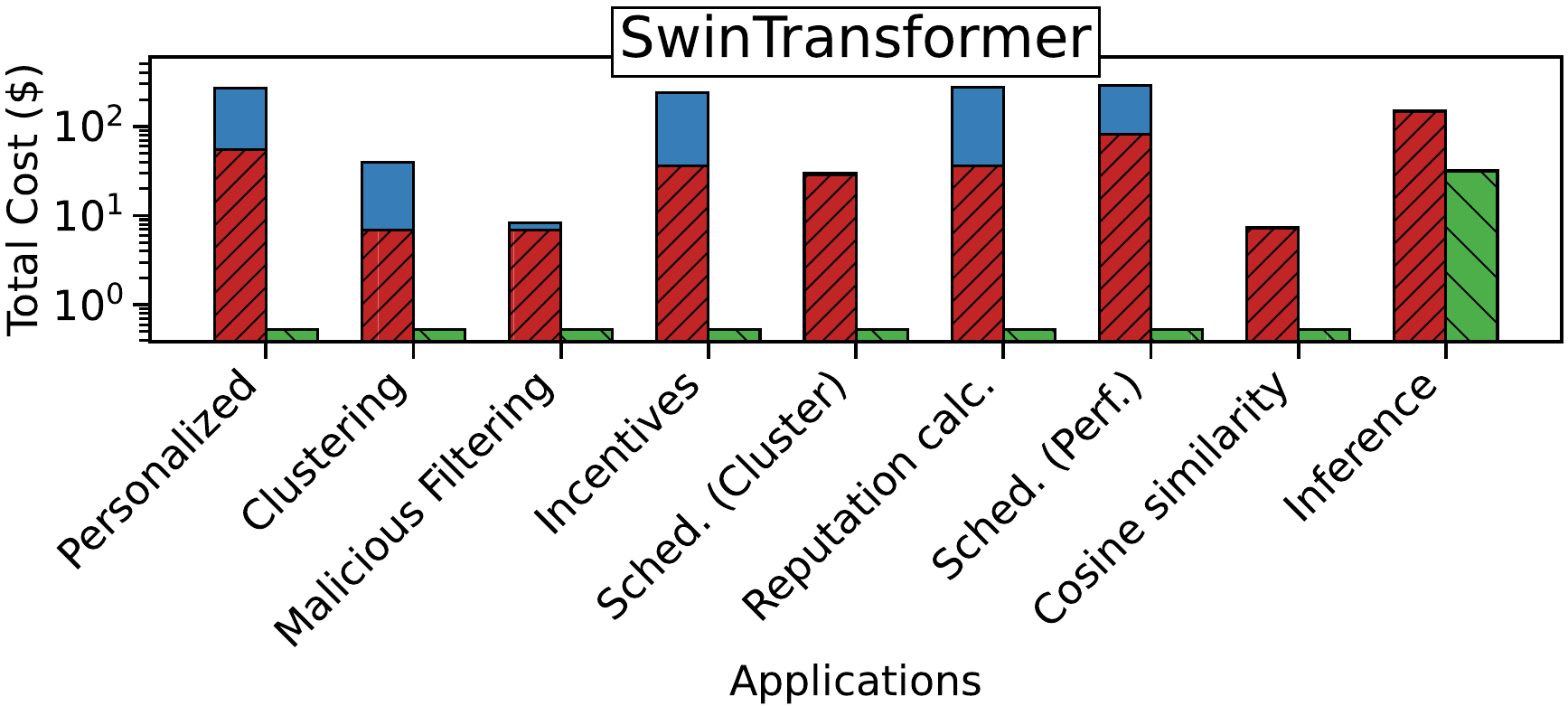}
   \vspace{-15pt}
   \label{subfig:FLStore_total_cost_breakup_SwinTransformer}
\end{subfigure}
  \caption{FLStore vs. ObjStore-Agg \textbf{total cost} breakup comparison over 50 hours.}
  \label{fig:total_cost_breakup_comparison}
  \vspace{-5pt}
\end{figure*}

\section{Latency and cost performance breakup}
\label{sec:total performance breakup}

To identify the bottleneck, we broke up the accumulated latency between communication and computation time over 50 hours of experiments for the 10 different workloads.

\subsection{FLStore vs ObjStore-Agg}
\label{subsec:FLStore vs ObjStore-Agg total latency breakup}

Figure~\ref{fig:total_time_breakup_comparison} shows the results with both communication and computation time for the ObjStore-Agg and only computation time for FLStore because communication time for FLStore is negligible in comparison due to co-located data and compute planes. The major bottleneck in ObjStore-Agg is Communication, in comparison the I/O time from memory to CPU is negligible~\cite{ninjaone_ddr_memory}.
For some workloads such as Inference, Debugging, and Scheduling, the difference between computation and communication times is significant. 
During inference communication consumes an average of $98.9\%$ of time. 
This shows that current methodologies for computing non-training workloads for distributed learning techniques such as FL are significantly communication-bound.
Thus, the reduction in communication times as brought by FLStore significantly improves the efficiency performance, which can be observed in Figure~\ref{fig:total_time_breakup_comparison}. 
We can also observe that FLStore provides significant improvements for smaller models, which is why FLStore is suitable in cross-device FL settings~\cite{kairouz2019advances,refl}. 
Across $50$ hours and $3000$ total requests we see Resnet18 with an average $82.04\%$ ($35.50$ second) decrease in latency,
MobileNet has an average $47.33\%$ ( $75.99$ second) decrease in latency,
EfficientNet has an average $50.44\%$ ($100.18$ second) decrease in latency, and
Swin has an average $20.45\%$ ($4.42$ second) decrease in latency.
Thus, FLStore can significantly improve non-training tasks in FL with reduced latency. We next observe the reduction in total cost with FLStore.

We perform the same breakup analysis on the costs in Figure~\ref{fig:total_cost_breakup_comparison}, showing both the communication and computation costs for ObjStore-Agg and computation costs for FLStore where communication costs are negligible. We can observe that the majority of the cost stems from the I/O (including communication) of data relevant to the non-training workloads. Resnet18, EfficientNet, and MobileNet spend $87.46\%$, $76.96\%$, and $85.80\%$ of their total time respectively in I/O, and SwinTransformer spends $53.32\%$ percent of its total time in I/O. Thus, by reducing the I/O time and data transfer costs FLStore provides a cost-effective solution for offloading the non-training workloads in FL.
Across 50 hours and 3000 total requests we see that Resnet18, MobileNet, and EfficientNet show a $94.73\%$, $92.72\%$, and $86.81\%$ average decrease in cost respectively, and
SwinTransformer has an average \(77.83\%\) reduction in cost.

\subsection{FLStore vs In-Memory Cache}
\label{subsec:FLStore vs ObjStore-Agg total latency breakup}

\begin{figure}
  \centering
{\includegraphics[width=0.55\columnwidth]{figures/elasticache_Legend_EfficientNet.pdf}}
\begin{subfigure}[t]{0.8\columnwidth}
  \centering
   \includegraphics[width=1\columnwidth]{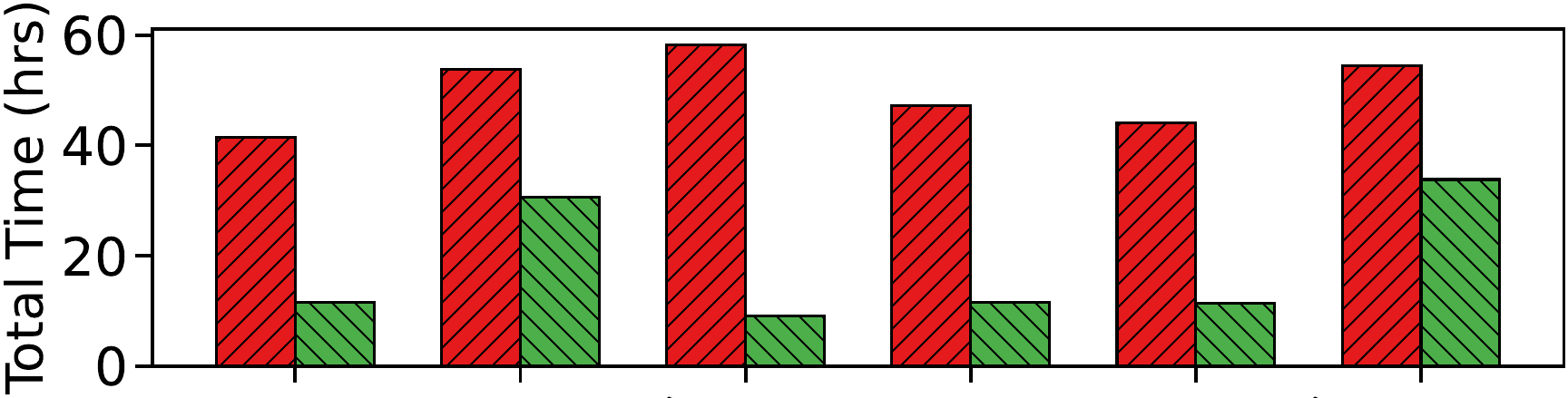}
   \label{subfig:ElastiCache_FLStore_total_time_summary}
\end{subfigure}
\begin{subfigure}[t]{0.82\columnwidth}
  \centering
   \includegraphics[width=1\columnwidth]{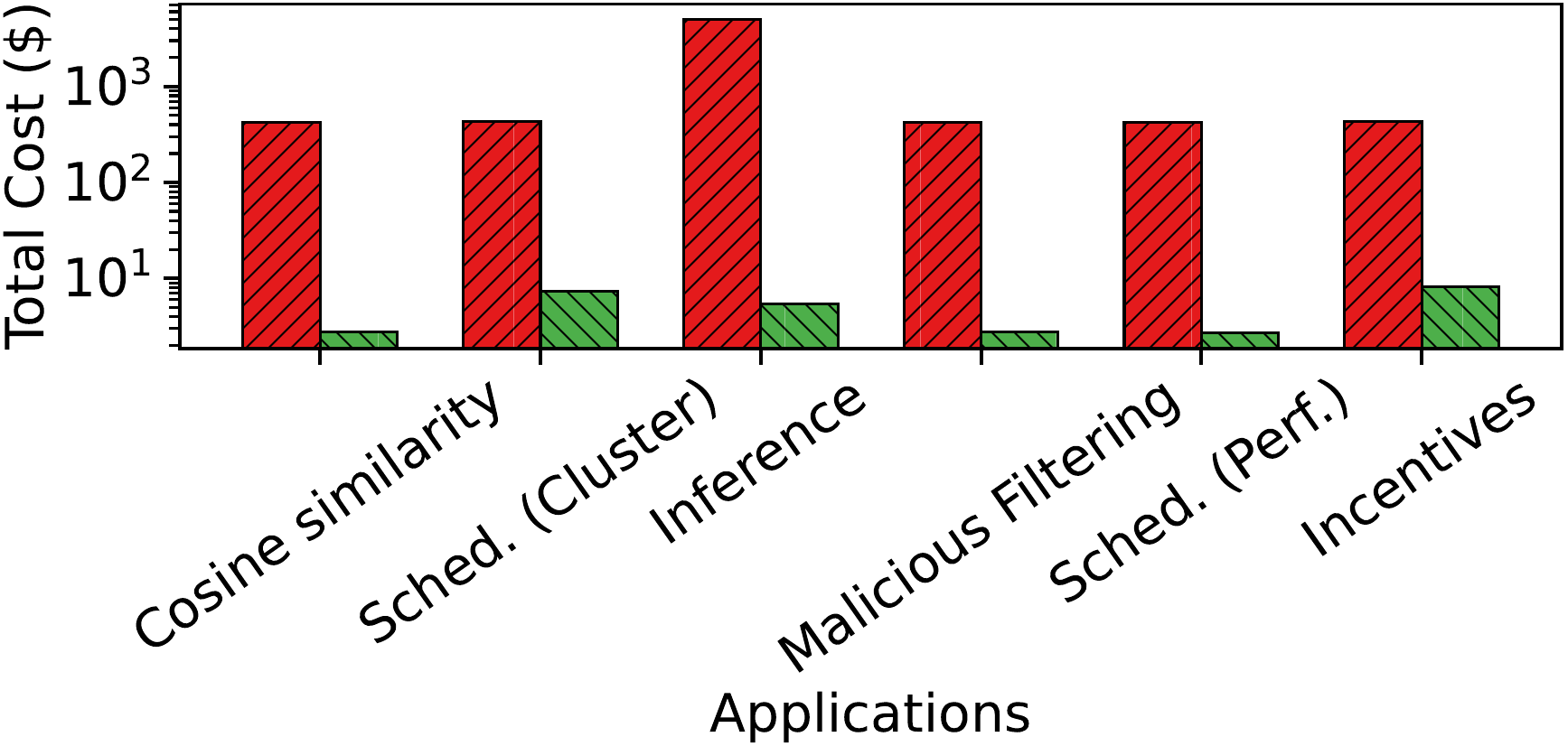}
   \label{subfig:elasticache_total_time_and_cost_comparison}
\end{subfigure}
\hspace{3pt}
  \caption{\textbf{Total time} (top) and \textbf{total cost} (bottom) comparison of Cache-Agg baseline vs. FLStore over 50 hours and 3000 total requests.}
  \label{fig:elasticache_total_time_and_cost_comparison}
\end{figure}

We also perform the total cost breakup analysis over 50 hours, 3000 total non-training requests, and 10 workloads, calculating both the communication and computation costs for Cache-Agg and FLStore. Results for this analysis are shown in Figure~\ref{fig:elasticache_total_time_and_cost_comparison} FLStore decreases the total time by $37.77\%-84.45\%$ amounting to 191.65 accumulated hours reduced for all requests and a $98.12\%-99.89\%$ decrease in total cost resulting in a reduction of $\$7047.16$ accumulated dollar costs for all 3000 requests across 50 hours. To compare both (Cache-Agg and ObjStore-Agg) on the same workloads tested with Cache-Agg, FLStore shows an average decrease in latency of $71\%$ with ObjStore-Agg and $64\%$ with Cache-Agg, the decreases with ObjStore-Agg is larger as cloud object stores are slower than cloud caches. However, in terms of costs cloud caches are more expensive than cloud object stores, which is why for the workloads tested with Cache-Agg, FLStore shows an average decrease in costs of $98.83\%$ compared to Cache-Agg and $92.45\%$ decrease compared to ObjStore-Agg.


\section{FLStore static: Ablation Study}
\label{sec:FLStore static: Ablation Study}
\begin{figure}[h]
\begin{subfigure}[t]{0.49\columnwidth}
  \centering
   \includegraphics[width=1\columnwidth]{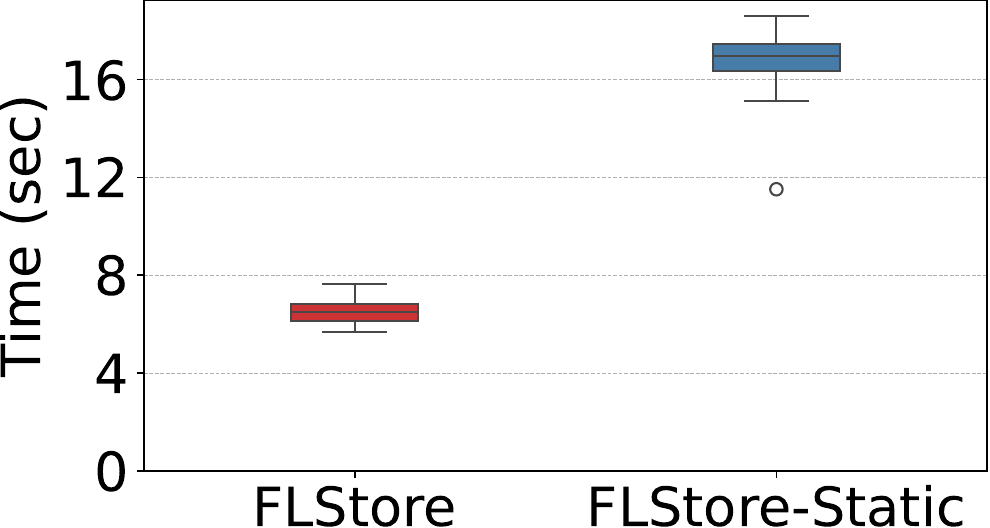}
\end{subfigure}
\begin{subfigure}[t]{0.5\columnwidth}
  \centering
   \includegraphics[width=1\columnwidth]{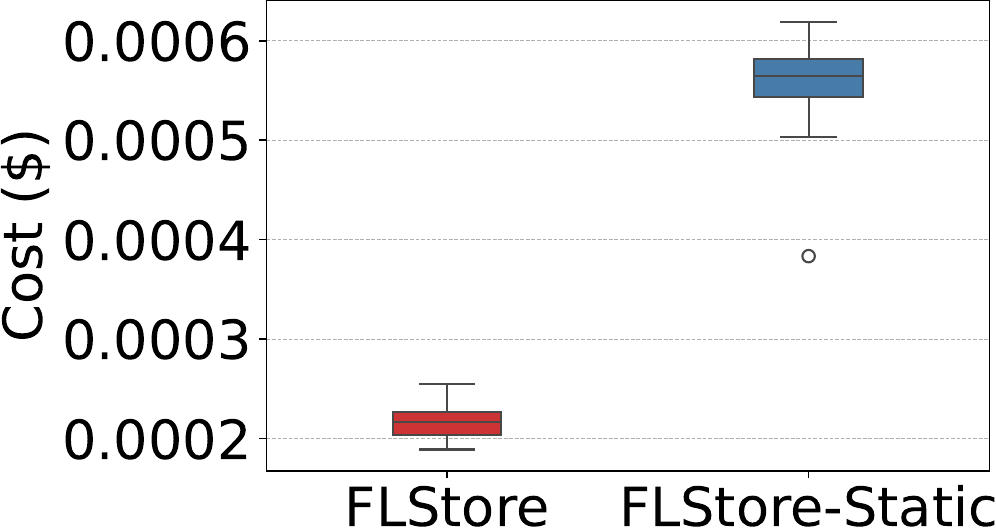}
\end{subfigure}
  \caption{FLStore vs. FLStore-Static: \textbf{Per request latency} (left) and \textbf{cost} (right) while filtering malicious clients.}
  \label{fig:FLStore_static_time_cost_comparison}
\end{figure}

For comparison with FLStore-Static, we consider a scenario where the workload changes from \textit{model inference} to \textit{malicious filtering}. Caching policy of FLStore-Static remains static (Individual Client Updates) which was for model inference workload while FLStore changes its caching policy to \textit{All Client Updates} based on the new workload (malicious filtering). Results in Figure~\ref{fig:FLStore_static_time_cost_comparison} show that FLStore reduces per-request average latency by $99\%$ (8 seconds) and costs by approximately $3\times$. This analysis highlights the importance of designing caching policies tailored for non-training FL workloads.

\section{Discussion: Limitations and Future Work}
\label{sec:Discussion}

\begin{figure}
  \centering
  \includegraphics[width=\linewidth]{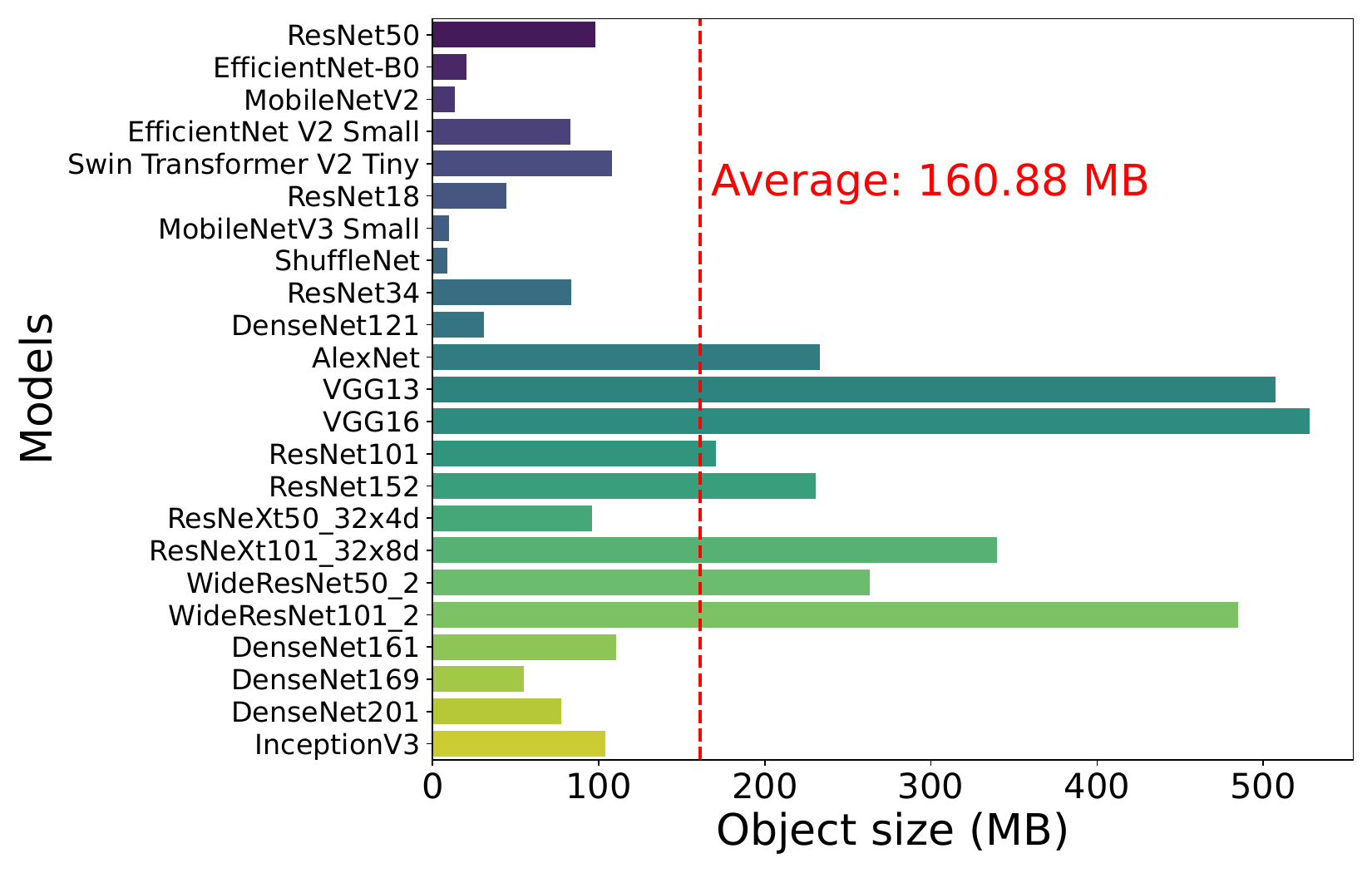}
  \caption{Memory footprint of commonly used models in FL.}
  \label{fig:model_footprint_per_request}
\end{figure}

\paragraph{Support for Foundation Models} Foundation Models are a class of models that have undergone training with a broad and general data set. Users can then fine-tune foundational models for specific use cases without training a model from scratch.
We have added and evaluated several foundation models from Figure~\ref{fig:model_footprint_per_request} in FLStore and continue to add more such models. We also add model inference as an application for FLStore with the aim of providing a cost-effective alternative for serving models efficiently compared to other cloud solutions such as AWS SageMaker~\cite{aws_sagemaker} which incurs high latency and costs as shown by our analysis in Figures~\ref{fig:total_time_breakup_comparison} and~\ref{fig:total_cost_breakup_comparison}.

\paragraph{FLStore Integration} Due to the modular nature of FLStore, it can be easily integrated into any existing FL framework. We have successfully integrated FLStore with IBMFL~\cite{ibmflgithub}, and FLOWER~\cite{beutel2020flower}, two popular FL frameworks used in industry and research.

\paragraph{Adaptive Caching Policies} Our ongoing efforts include designing agents based on Reinforcement Learning with Human Feedback (RLHF) that can understand the characteristics of non-training workloads and create new caching policies for those workloads using our existing caching policies as a base. RLHF has successfully been deployed for hyperparameter and optimization configuration in FL~\cite{Khan2024FLOAT} and the configuration of caching policies is a similar challenge that we hope to resolve by employing this technique.

\paragraph{Function Memory Limitations} Serverless functions are limited in memory resources having a maximum of 10 GB memory~\cite{aws_lambda}. This is more than sufficient for handling non-training workloads for cross-device FL even for small transformer models such as~\cite{TinyLlama2024}. As shown in Figure~\ref{fig:model_footprint_per_request}, the average size of popular models used in cross-device FL is just 161 MB approximately. However, for even larger foundational models such as Large Language Models (LLMs)~\cite{NEURIPS2020_gpt3}, we are working on utilizing pipeline parallel processing where function groups can be assigned for each workload and each function in that group can perform computations in a pipeline parallel manner~\cite{serverlessML,INFless}.

\end{document}